\documentclass[times,twocolumn,final,authoryear]{elsarticle}
\usepackage{amsmath}
\usepackage{algorithm} 
\usepackage{algorithmic}

\def\etal{{\em et al.}}
\def\eg{{\em e.g.}}
\def\ie{{\em i.e.}}

\emergencystretch=\hsize
\graphicspath{{doc/}}

\usepackage{booktabs}  
\usepackage{multirow}
\usepackage{bm}
\usepackage{array}
\usepackage{graphicx}
\usepackage{epstopdf}
\usepackage{subfigure}
\usepackage{amssymb}
\usepackage{amsmath}
\usepackage{makecell}
\usepackage{framed,multirow}
\usepackage{amssymb}
\usepackage{latexsym}
\usepackage{url}
\usepackage{xcolor}
\usepackage{hyperref}

\usepackage{soul}

\sethlcolor{yellow}

\definecolor{newcolor}{rgb}{.8,.349,.1}

\biboptions{authoryear}

\begin{document}


\makeatletter
\def\ps@pprintTitle{%
  \let\@oddhead\@empty
  \let\@evenhead\@empty
  \let\@oddfoot\@empty
  \let\@evenfoot\@empty
}
\makeatother

\begin{frontmatter}

\title{Noise-Aware Boundary-Enhanced Generative Learning for Ultrasound Speckle Reduction}

\author[1]{Yuexi {Gu}}

\author[1,2]{Mengqi {Wu}}

\author[1]{Yongheng {Sun}}

\author[1]{Virginie {Papadopoulou}}

\author[1]{Mingxia {Liu}\corref{cor1}}

\author[1]{Maureen {Kohi}\corref{cor1}}

\cortext[cor1]{Corresponding authors: M.~Liu (Email: mingxia\_liu@med.unc.edu) and M.~Kohi (Email: maureen\_kohi@med.unc.edu).}

\address[1]{Department of Radiology and Biomedical Research Imaging Center (BRIC), University of North Carolina at Chapel Hill, Chapel Hill, NC 27599, USA}
\address[2]{Lampe Joint Department of Biomedical Engineering, University of North Carolina at Chapel Hill, Chapel Hill, NC 27599, USA.}

\begin{abstract}
Ultrasound is a non-invasive, real-time, and cost-effective imaging technique widely used in clinical diagnosis. However, its diagnostic efficacy is often compromised by \emph{inherent speckle noise} that degrades image quality and obscures underlying anatomical structures. Existing speckle reduction methods tend to over-smooth tissue boundaries and generalize poorly to heterogeneous noise levels. To address these limitations, we propose a Noise-Aware Boundary-Enhanced Generative Learning (\textbf{NBGL}) framework for ultrasound speckle reduction, which simultaneously preserves annotated anatomical boundaries and adapts to varying noise levels. The NBGL framework consists of a \emph{speckle reduction branch} and a \emph{boundary enhancement branch}. 
The former leverages generative learning to suppress speckle noise, while the latter learns boundary-sensitive representations to preserve target anatomical structures. Furthermore, a \emph{noise-aware interaction weight generation (NIWG) module} estimates the speckle noise level via 3D Laplacian filtering and a median absolute deviation estimator, and translates it into an adaptive interaction weight. This weight is incorporated into a \emph{weighted feature-wise linear modulation (wFiLM) module} to adaptively modulate cross-branch feature coupling, thereby improving robustness to varying noise levels. Extensive evaluations on 141 3D transvaginal ultrasound volumes demonstrate that NBGL consistently outperforms state-of-the-art methods in speckle reduction and structural preservation across six noise levels, while maintaining consistency with annotated anatomical boundaries.

\end{abstract}


\end{frontmatter}

\section{Introduction}
\label{sec:introduction}
Ultrasound imaging is widely utilized in clinical diagnosis because it is non-invasive, radiation-free, real-time, and cost-effective. 
However, its image quality is often degraded by inherent speckle, which appears as a granular pattern~\citep{burckhardt2005speckle}. 
Speckle arises from the interference of echoes scattered by sub-resolution tissue structures and is 
generally modeled as \emph{multiplicative noise}, where the observed ultrasound intensity is formed by multiplying the underlying speckle-free intensity by a random speckle component. 
The fluctuation of this random component reflects the speckle noise level~\citep{wagner2005statistics, michailovich2006despeckling}, which can vary across ultrasound volumes due to differences in acquisition settings, imaging depth, and tissue characteristics.
Stronger speckle can obscure subtle anatomical details and weaken tissue boundary contrast, affecting accurate clinical assessment.
Therefore, ultrasound speckle reduction, also referred to as \emph{despeckling}, remains a challenging task that requires effective speckle suppression while preserving anatomical structures under heterogeneous noise levels.

\begin{figure*}[!t]
\setlength{\abovecaptionskip}{0pt}
\setlength{\belowcaptionskip}{0pt}
\setlength{\abovedisplayskip}{0pt}
\setlength{\belowdisplayskip}{0pt}
\centering
\includegraphics[width=1\textwidth]{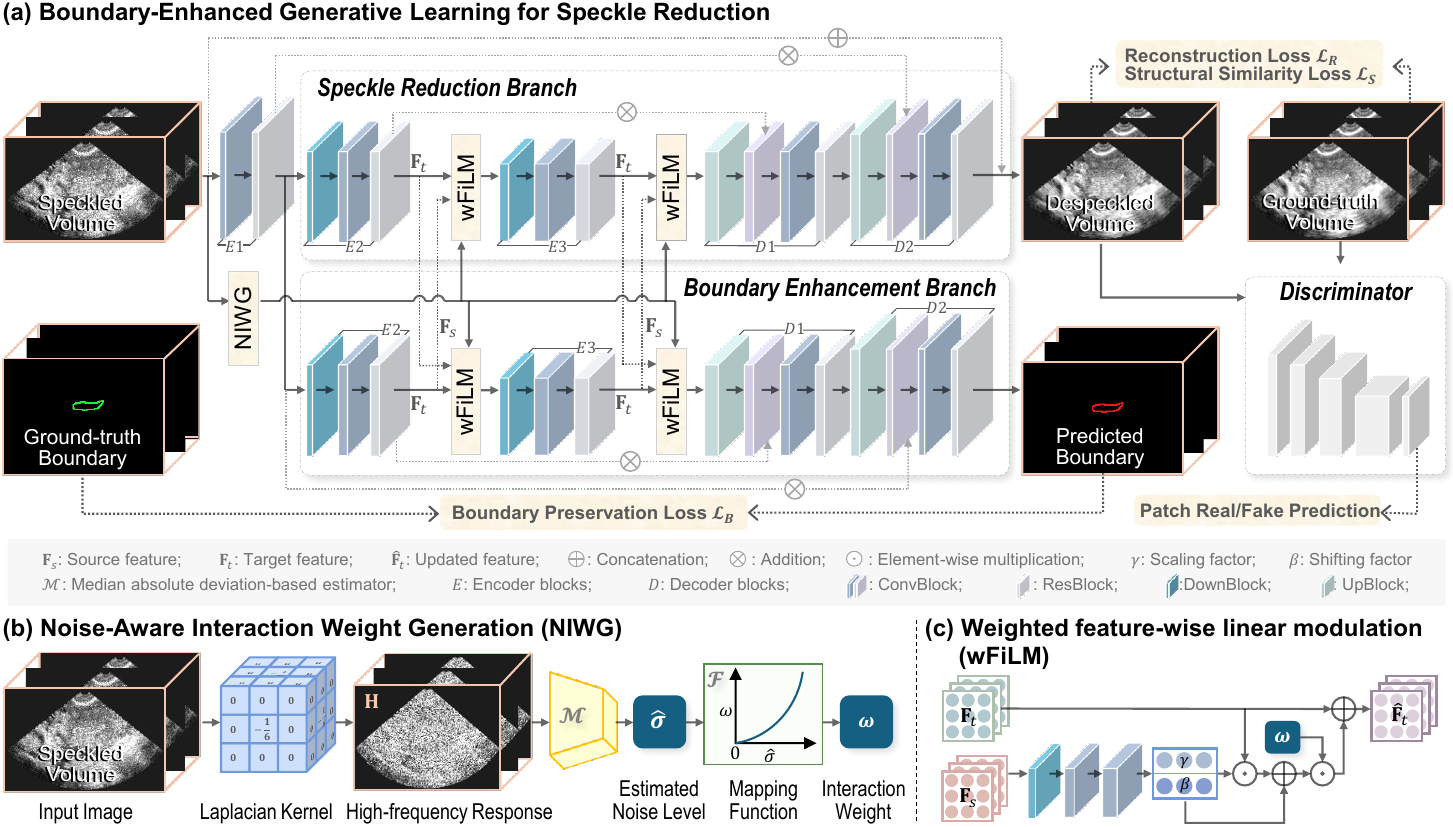}
    \caption{Overview of the proposed noise-aware boundary-enhanced generative learning (NBGL) framework for ultrasound speckle reduction. 
    (a) \emph{Boundary-enhanced generative learning}, consisting of a dual-branch generator and a patch discriminator. In the generator, a speckle reduction branch and a boundary enhancement branch are adaptively coupled to preserve anatomical boundaries, while the patch discriminator estimates patch-level real/fake scores for adversarial training, encouraging realistic anatomical textures. Cross-branch feature coupling is adaptively modulated by the wFiLM module via an interaction weight $\omega$ computed from the estimated noise level.  
    (b) \emph{Noise-aware interaction weight generation} (NIWG) module. The speckle noise level $\hat{\sigma}$ is estimated by applying a median absolute deviation estimator to the high-frequency response obtained from 3D Laplacian filtering, and is then mapped to the interaction weight $\omega$ for wFiLM-based cross-branch coupling.
    (c) \emph{Weighted feature-wise linear modulation} (wFiLM) module. The source feature $\mathbf{F}_s$ generates affine parameters that modulate the target feature $\mathbf{F}_t$, with the modulation strength scaled by $\omega$.}
    \label{fig:NBGL_pipeline}
\end{figure*}

Existing approaches for ultrasound speckle reduction can be broadly grouped into \emph{filtering-based methods} and \emph{deep learning-based methods}. 
Filtering-based methods, such as bilateral filtering~\citep{tomasi1998bilateral} and BM4D~\citep{maggioni2012nonlocal}, reduce speckle through local smoothing or non-local patch grouping, respectively. 
Although these methods are usually training-free and computationally efficient, their manually specified filtering rules or similarity assumptions may blur tissue boundaries and offer limited adaptability to heterogeneous noise levels~\citep{loizou2005comparative, joel2018extensive}. 
In contrast, deep learning-based methods provide a more flexible framework by learning speckle reduction models from ultrasound data~\citep{kokil2020despeckling}. However, many of them are primarily optimized for intensity restoration, which can still result in over-smoothed tissue boundaries and unstable performance across varying noise levels~\citep{nazir2024recent}.  
To better preserve structural details, several studies have introduced gradient- or edge-based constraints into speckle reduction models~\citep{zhou2024gradient, han2022dual, qiu2023despeckling}. 
While these methods improve boundary awareness, they often rely on image-level gradient or edge representations. 
Such representations may become unreliable as the speckle noise level increases, because speckle-induced high-frequency components can be difficult to distinguish from anatomical boundaries. 
As a result, preserving anatomically meaningful boundaries remains challenging when structural guidance is extracted directly from speckled ultrasound images. 
In parallel, heterogeneous speckle levels in clinical ultrasound volumes introduce an additional robustness challenge for speckle reduction models. 
Models trained under a specific noise level may not generalize well to inputs with different speckle levels. To address this issue, noise-adaptive methods estimate the input noise level or train models under multiple noise levels to improve robustness~\citep{lan2020real, sudharson2021noise}, but they generally do not incorporate explicit anatomical boundary guidance for preserving clinically relevant boundaries. 
These limitations highlight the need for an ultrasound speckle reduction framework that can preserve clinically relevant anatomical boundaries while adapting to varying speckle levels.

To this end, this study proposes a Noise-Aware Boundary-Enhanced Generative Learning (NBGL) framework for ultrasound speckle reduction.
As illustrated in Fig.~\ref{fig:NBGL_pipeline}, NBGL adopts a dual-branch architecture for joint speckle reduction and boundary enhancement. 
The speckle reduction branch generates despeckled volumes under adversarial supervision from a patch-level discriminator, while the boundary enhancement branch uses annotation-derived anatomical boundaries
as explicit geometric guidance to prevent the over-smoothing of fine structural edges.
A noise-aware interaction weight generation (NIWG) module estimates the speckle noise level by applying a median absolute deviation (MAD) estimator to the high-frequency response obtained from 3D Laplacian filtering, and maps the estimate through a nonlinear function to an interaction weight. 
This weight is incorporated into the weighted feature-wise linear modulation (wFiLM) module, which scales the bidirectional feature coupling between the speckle reduction and boundary enhancement branches according to the estimated noise level.
The framework is trained end-to-end with a speckle reduction loss combining reconstruction and structural similarity terms, a boundary preservation loss, and an adversarial loss. 

The main contributions of this work are summarized below:
\begin{itemize}
    \item We propose a novel Noise-Aware Boundary-Enhanced Generative Learning (NBGL) framework that, for the first time, seamlessly integrates adversarial speckle reduction with explicit annotation-guided boundary enhancement in a unified 3D architecture. By establishing a dual-branch learning scheme, our framework fundamentally mitigates the chronic over-smoothing of delicate anatomical boundaries commonly suffered by existing despeckling models. 
    \item We introduce an innovative adaptive feature coupling mechanism driven by two newly designed components: a Noise-Aware Interaction Weight Generation (NIWG) module and a Weighted Feature-Wise Linear Modulation (wFiLM) module. 
    They achieve dynamic, bidirectional cross-branch feature modulation by accurately estimating voxel-level speckle noise via 3D Laplacian filtering and MAD, thereby yielding exceptional robustness against highly heterogeneous and volatile noise distributions.
    \item Extensive validation on 141 3D transvaginal ultrasound volumes from the UterUS dataset~\citep{bonevs2024automatic} demonstrates that NBGL consistently outperforms the state-of-the-art filtering and deep learning-based benchmarks. Quantitative and qualitative evaluations across six distinct noise levels verify our framework's superior capability in simultaneous intensity restoration, clinical structure preservation, and strict fidelity to annotated anatomical boundaries. 
\end{itemize}

The remainder of this paper is organized as follows. Section~\ref{sec:related_work} reviews related work on ultrasound speckle reduction. 
Section~\ref{sec:method} introduces the proposed framework. 
Section~\ref{sec:experiments} presents quantitative and qualitative evaluations across multiple noise levels. 
Section~\ref{sec:discussion} analyzes the influence of several key components of the proposed method and outlines its limitations and future directions. Section~\ref{sec:conclusion} concludes the paper.

\section{Related Work}
\label{sec:related_work}
\subsection{Filtering-Based Speckle Reduction}
Filtering-based speckle reduction methods can be broadly categorized into local spatial-domain filters, adaptive local statistical filters, and non-local transform-domain methods~\citep{michailovich2006despeckling, joel2018extensive}. 
Local spatial-domain filters, including Gaussian and mean filtering~\citep{jahne2005digital}, median filtering~\citep{huang1979fast}, and bilateral filtering~\citep{tomasi1998bilateral}, directly operate on local neighborhoods with predefined filtering rules. 
In particular, Gaussian and mean filters reduce noise by averaging nearby intensities, whereas median filtering relies on local order statistics and is therefore more robust to outliers such as high-intensity scatterers. Bilateral filtering further combines spatial and intensity-similarity weighting, thereby smoothing nearby samples with similar intensities while reducing averaging across intensity edges. 

Although these filters are computationally simple, their predefined local operations may blur tissue boundaries together with speckle and weaken fine anatomical details. 
To address this, 
adaptive local statistical filters such as Lee~\citep{lee1980digital}, Frost~\citep{frost1982model}, and Wiener~\citep{lim1990two} compute each output pixel as a weighted combination of the original intensity and the local mean, with weights determined by local statistics such as variance. These filters apply stronger smoothing in relatively homogeneous regions and weaker smoothing in regions with higher local variation, helping preserve anatomical structures. 
However, these filters still rely on statistics estimated from a fixed local window. 
Under severe speckle noise, such local statistics can become unreliable, and the fixed window also prevents the use of structurally similar regions outside the local neighborhood.

To address this locality constraint, 
non-local transform-domain methods are proposed to reduce speckle by searching for similar patches across the volume. 
For example, the Block-Matching and 4D filtering (BM4D) algorithm~\citep{maggioni2012nonlocal} groups similar 3D patches into a higher-dimensional stack and performs collaborative filtering in a transform domain. Since the grouped patches contain similar anatomical or textural structures but different speckle patterns, collaborative filtering can suppress inconsistent speckle components while preserving structural details. 
In summary, filtering-based methods are training-free, but they often over-smooth anatomical details, especially under severe speckle noise, and lack the flexibility to handle varying noise levels in 3D ultrasound volumes.

\subsection{Deep Learning-Based Speckle Reduction}
Compared with filtering-based methods, deep learning-based speckle reduction models learn data-driven representations or image priors from ultrasound data, providing a more flexible framework for modeling complex speckle patterns and anatomical structures during despeckling.
Existing deep learning-based methods are based on a wide range of network designs. 
Convolutional and transformer-based networks~\citep{sivaanpu2024speckle, chen2024lit} learn speckle reduction mappings from noisy inputs. 
Adversarial learning frameworks~\citep{jimenez2024ultrasound, sikhakhane2024review, huang2021gan, han2022dual} introduce discriminator-based supervision to improve the realism of despeckled outputs and preserve fine structural details. 
Generative models, such as variational autoencoders~\citep{rombach2022high} and diffusion-based methods~\citep{asgariandehkordi2023deep, kim2025tackling}, formulate speckle reduction as a generative or iterative restoration process. 
In parallel, self-supervised approaches including Noise2Noise and its variants~\citep{lehtinen2018noise2noise, li2025speckle2self, gobl2022speckle2speckle} learn from noisy observations alone via statistical assumptions about the speckle distribution. Despite these architectural advances, most existing models prioritize intensity-level speckle reduction and still struggle to simultaneously preserve fine anatomical boundaries and remain robust across varying noise levels~\citep{nazir2024recent, kim2024systematic}.

To mitigate over-smoothing issues, several studies have introduced structure-aware constraints into speckle reduction models. 
DGGNet~\citep{wang2024dggnets} explicitly preserves low-level gradient information using a dual-decoder design, where one decoder recovers the despeckled image, and the other constrains gradient consistency to retain structural details. 
Extending this idea to adversarial learning, DU-GAN~\citep{huang2021gan} introduces dual-domain discriminators in the image and gradient domains, using gradient-domain supervision to enhance edge-related structures. 
Beyond gradient-based constraints, CAPAD~\citep{qiu2023despeckling} uses an attention-fusion encoding module to extract image features from ultrasound images and a Canny edge-based joint loss to preserve structural details during speckle reduction. However, these methods mainly derive structural guidance from the speckled input or its gradient representation. 
Under severe speckle noise, image-level gradient or edge responses may include speckle-induced high-frequency components, reducing the reliability of such structure-aware guidance.
Moreover, these methods are typically trained under fixed noise conditions, limiting their effectiveness when the input noise level changes.

To improve robustness to varying noise levels, several studies have incorporated noise-adaptive strategies into speckle reduction models. 
Kokil and Sudharson~\citep{kokil2020despeckling} train a residual learning network using reference images corrupted by simulated speckle noise at varying levels, improving its robustness under noisy input conditions. 
Lan and Zhang~\citep{lan2020real} propose a mixed-attention residual U-Net for real-time ultrasound speckle reduction, where speckle noise levels are graded at fixed intervals and estimated for real ultrasound images to guide speckle reduction. 
Sudharson~\etal~~\citep{sudharson2021noise} propose a noise-level estimation method that extracts noise-aware features from high-pass filtered ultrasound images and uses support vector regression to estimate the unknown speckle noise levels. 
These methods improve robustness by training models under different noise levels or by estimating the input noise level. 
However, these methods focus primarily on speckle noise adaptation while neglecting explicit geometric guidance. Consequently, existing studies rarely integrate noise-level estimation with boundary preservation within a unified framework, thereby limiting their capability to jointly preserve critical structural boundaries and maintain robustness across varying noise levels in ultrasound speckle reduction. 

\section{Proposed Method}
\label{sec:method}
\subsection{Overall Framework} 
Let $\mathbf{X}, \mathbf{Y} \in \mathbb{R}^{H \times W \times D}$ denote a speckled ultrasound volume and its corresponding ground-truth (GT) volume, respectively. 
Let $\mathbf{B} \in \{0,1\}^{H \times W \times D}$ denote the boundary of a target anatomical structure (\eg,  uterine cavity in this work). Given the speckled volume $\mathbf{X}$ as input, the proposed framework estimates a despeckled volume $\hat{\mathbf{Y}}$ and predicts an auxiliary boundary probability map $\hat{\mathbf{B}}$. The GT boundary map $\mathbf{B}$ is used only to supervise $\hat{\mathbf{B}}$ during training and is not required at inference time.

As illustrated in Fig.~\ref{fig:NBGL_pipeline}(a), the proposed NBGL is formulated as a boundary-enhanced generative learning framework for ultrasound speckle reduction. 
The framework contains a \emph{dual-branch generator} and a \emph{discriminator}. 
The generator consists of a shared encoder block $E1$ followed by a speckle reduction branch and a boundary enhancement branch, which generate $\hat{\mathbf{Y}}$ and $\hat{\mathbf{B}}$, respectively. 
The discriminator provides adversarial supervision for the despeckled output. 
Within the generator, cross-branch feature coupling is controlled by the noise-aware interaction weight generation (NIWG) module in Fig.~\ref{fig:NBGL_pipeline}(b) and the weighted feature-wise linear modulation (wFiLM) module in Fig.~\ref{fig:NBGL_pipeline}(c). 
The framework is trained end-to-end with speckle reduction, boundary preservation, and adversarial objectives.

\subsection{Boundary-Enhanced Generative Learning}
As shown in Fig.~\ref{fig:NBGL_pipeline}(a), boundary-enhanced generative learning consists of a dual-branch generator and a discriminator.

\textbf{Dual-Branch Generator.} 
The generator follows an encoder-decoder design with a shared feature extractor, two branch-specific encoders, and two parallel decoders. The input volume $\mathbf{X}$ is first processed by the shared encoder block $E1$ to obtain low-level volumetric features, which are then fed into both the speckle reduction and the boundary enhancement branches. 
Each branch contains its own encoder blocks, denoted as $E2$ and $E3$, where feature maps are downsampled, and channel dimensionality is increased. 
The speckle reduction encoder extracts features for reducing speckle noise, whereas the boundary enhancement encoder extracts boundary-sensitive features guided by the annotation-derived boundary of target anatomical structures.

After each selected encoder stage $l \in \{2,3\}$, the corresponding branch features are coupled by a stage-specific wFiLM module, with the interaction weight generated by the NIWG module. 
The resulting latent features in each branch are then processed by its own decoder blocks, denoted as $D1$ and $D2$, where feature maps are upsampled and combined with the corresponding encoder features through skip connections. 
The speckle reduction decoder predicts a residual term $\mathbf{R}$, yielding the despeckled volume $\hat{\mathbf{Y}}=\mathbf{X}+\mathbf{R}$. 
The boundary enhancement decoder predicts the boundary probability map $\hat{\mathbf{B}}$.

\textbf{Discriminator.} A 3D patch-level discriminator is then designed to provide adversarial supervision for the despeckled output. 
Specifically, it takes paired volumes $(\mathbf{X}, \mathbf{Y})$ and $(\mathbf{X}, \hat{\mathbf{Y}})$ as real and generated pairs, respectively, where $\mathbf{X}$ is the speckled volume, $\mathbf{Y}$ is the GT volume, and $\hat{\mathbf{Y}}$ is the generated despeckled volume. 
By learning to distinguish these real pairs from the generated ones, the discriminator enforces a conditional distribution alignment between the synthesized and ground-truth volumes. 
Through this adversarial supervision, the generated volume is encouraged to retain local anatomical details and texture consistency.

\subsection{Noise-Aware Interaction Weight Generation (NIWG)}
\label{sec:NIWG}

The cross-branch feature coupling in NBGL is fundamentally bidirectional, enabling mutual feature modulation between the boundary enhancement and speckle reduction branches.
This coupling strength must dynamically adapt to the input's speckle noise level. 
Specifically, at elevated noise levels, intense speckle patterns increasingly obscure anatomical boundaries, making them heavily indistinguishable. Under such degraded conditions, anatomical boundary-aware features become indispensable for guiding boundary-preserving speckle reduction; conversely, features from the speckle reduction branch offer cleaner, denoised representations that facilitate more accurate boundary feature extraction. Motivated by this complementarity, as illustrated in Fig.~\ref{fig:NBGL_pipeline}(b), the proposed NIWG module is designed to estimate the speckle noise level directly from the input volume and convert it into an adaptive interaction weight, which is then incorporated into the wFiLM module to precisely regulate inter-branch coupling strength.

\subsubsection{Speckle Noise Level Estimation}
In ultrasound images, speckle visually appears as a granular pattern, while at the intensity level it corresponds to rapid local intensity changes that are reflected in high-frequency responses~\citep{pizurica2003versatile, sudha2009speckle}. 
Therefore, the speckle noise level can be estimated from the magnitude of high-frequency responses. 
To extract this high-frequency response, we apply a discrete 3D Laplacian kernel $\mathbf{K}_{L}$
to the input volume. 
As a zero-sum second-order operator, the Laplacian gives near-zero responses in smooth, slowly varying regions and larger responses where the intensity changes rapidly~\citep{immerkaer1996fast, tai2008fast}. 
Specifically, for an input volume $\mathbf{X}$, the high-frequency response $\mathbf{H}$ is computed as:
\begin{equation}
    \mathbf{H} = \mathbf{X} * \mathbf{K}_{L},
\end{equation}
where $*$ denotes 3D convolution, and $\mathbf{K}_{L}$ is a normalized $3\times3\times3$ Laplacian kernel with weights $1$ at the center, $-\frac{1}{6}$ at the six axial neighbors, and $0$ elsewhere.

Although $\mathbf{H}$ emphasizes high-frequency responses, it may also contain responses from anatomical structures such as tissue boundaries. 
Therefore, directly estimating the noise level from all values in $\mathbf{H}$ may be biased by these structural high-frequency responses. 
To obtain a robust estimate of the typical high-frequency magnitude, we apply a median absolute deviation (MAD)-based estimator~\citep{donoho1994ideal} 
to $\mathbf{H}$ to obtain the estimated speckle noise level $\hat{\sigma}$:
\begin{equation}
    \hat{\sigma} = c_{\mathrm{MAD}} \cdot 
    \mathrm{median}\left(
    \left|\mathbf{H} - \mathrm{median}(\mathbf{H})\right|
    \right),
 \label{eq:mad}
\end{equation}
where $c_{\mathrm{MAD}}=1.4826$ is the standard Gaussian calibration constant. 
The MAD estimator yields a robust approximation of the input speckle intensity while effectively filtering out structural high-frequency outliers, making it superior to standard variance estimators in complex tissue regions. This parameter-free estimation ensures highly stable and consistent interaction weight generation during both training and inference.

\subsubsection{Interaction-Weight Generation}
This estimate $\hat{\sigma}$ is then converted into an interaction weight that modulates the strength of cross-branch feature coupling. 
Since $\hat{\sigma}$ may vary across input volumes and intensity scales, it is first normalized to obtain $\tilde{\sigma} \in [0,1]$ with respect to the expected speckle noise range. 
As higher speckle noise levels can more severely obscure tissue boundaries, a higher estimated noise level corresponds to a larger interaction weight. 
The normalized noise level $\tilde{\sigma}$ is then mapped to an interaction weight through a bounded power-law function:
\begin{equation}
    \omega = \omega_{\min} + (\omega_{\max} - \omega_{\min}) \cdot \tilde{\sigma}^{\rho},
\label{eq:interaction_weight_mapping}
\end{equation}
where $\omega_{\min}$ and $\omega_{\max}$ are predefined hyperparameters specifying the lower and upper bounds of the interaction weight, respectively, and $\rho$ controls the nonlinearity of the mapping. When $\rho>1$, the interaction weight changes slowly at lower noise levels and increases more rapidly as the estimated noise level approaches the upper range. The bounded form ensures that the cross-branch modulation strength remains within a stable range throughout training and inference, with the values of $\omega_{\min}$, $\omega_{\max}$, and $\rho$ specified in Section~\ref{sec:implementation}.

Based on the interaction weight $\omega$, direction-specific weights are further defined for bidirectional coupling between the two branches. Let $\mathrm{r}$ and $\mathrm{b}$ index the speckle reduction and boundary enhancement branches, respectively. 
As shown in Fig.~\ref{fig:NBGL_pipeline}(a), wFiLM is applied after the second and third encoder stages, denoted as interaction stages $l \in \{2,3\}$. 
At each interaction stage, the base weight $\omega$ is assigned to $\omega_{\mathrm{r}\leftarrow\mathrm{b}}^{(l)}$, while the reverse direction $\omega_{\mathrm{b}\leftarrow\mathrm{r}}^{(l)}$ is scaled by $\alpha^{(l)}$:
\begin{equation}
    \omega_{\mathrm{r}\leftarrow\mathrm{b}}^{(l)} = \omega, \quad
    \omega_{\mathrm{b}\leftarrow\mathrm{r}}^{(l)} = \alpha^{(l)} \omega,
\label{eq:dynamic_interaction_weights}
\end{equation}
where $\alpha^{(l)}$ is a direction-specific attenuation factor. 
Specifically, at the earlier interaction stage, $\alpha^{(2)} = 1$ enforces a symmetric feature coupling to facilitate rich cross-branch representation sharing. At the deeper interaction stage, setting $\alpha^{(3)} = 0.2$ strategically suppresses the modulation flow from the speckle reduction branch to the boundary enhancement branch. This asymmetric design mitigates the risk of propagating over-smoothed intensity representations (often resident in deep denoising features) into the boundary stream, thereby safeguarding fine and delicate anatomical boundary structures.

\subsection{Weighted Feature-Wise Linear Modulation (wFiLM)}
\label{sec:wFiLM}
Given the direction-specific interaction weights defined above, the proposed wFiLM module extends standard FiLM~\citep{perez2018film} by introducing noise-aware scaling into cross-branch modulation. This module enables bidirectional feature coupling between the speckle reduction and boundary enhancement branches. 
Through this coupling, boundary-sensitive features guide boundary-preserving speckle reduction, while features from the speckle reduction branch provide cleaner representations for distinguishing anatomical boundaries from speckle patterns in the boundary enhancement branch.

Specifically, at interaction stage $l$, the update of the target feature $\mathbf{F}_{t}^{(l)}$ in branch $t$ is conditioned on the source feature $\mathbf{F}_{s}^{(l)}$ from branch $s$ and scaled by the noise-aware interaction weight $\omega_{t \leftarrow s}^{(l)}$, where $s,t \in \{\mathrm{r}, \mathrm{b}\}$ and $s \neq t$. 
First, to condition branch $t$ on branch $s$, $\mathbf{F}_{s}^{(l)}$ is mapped to channel-wise affine parameters:
\begin{equation}
    \left(
    \boldsymbol{\gamma}_{t \leftarrow s}^{(l)}, 
    \boldsymbol{\beta}_{t \leftarrow s}^{(l)}
    \right)
    =
    \Phi_{t \leftarrow s}^{(l)}
    \left(\mathbf{F}_{s}^{(l)}\right),
\end{equation}
where $\Phi_{t \leftarrow s}^{(l)}$ is a stage-specific and direction-specific parameter generation function, while $\boldsymbol{\gamma}_{t \leftarrow s}^{(l)}$ and $\boldsymbol{\beta}_{t \leftarrow s}^{(l)}$ denote the channel-wise scaling and shifting coefficients, respectively. 
Given these affine parameters, a residual modulation term is then generated for the target feature: 
\begin{equation}
    \Delta \mathbf{F}_{t \leftarrow s}^{(l)}
    =
    \boldsymbol{\gamma}_{t \leftarrow s}^{(l)}
    \odot \mathbf{F}_{t}^{(l)}
    +
    \boldsymbol{\beta}_{t \leftarrow s}^{(l)},
\end{equation}
where $\odot$ denotes channel-wise multiplication. 
Finally, the target feature is updated by adding the residual modulation term, whose contribution is controlled by the interaction weight:
\begin{equation}
    \hat{\mathbf{F}}_{t}^{(l)}
    =
    \mathbf{F}_{t}^{(l)}
    +
    \omega_{t \leftarrow s}^{(l)}
    \Delta \mathbf{F}_{t \leftarrow s}^{(l)} .
\label{eq:film_feature_update}
\end{equation}
Here, $\omega_{t \leftarrow s}^{(l)}$ is determined by the estimated speckle noise level and scales the cross-branch modulation strength. 
Applying this update in both directions using the weights defined in Eq.~\eqref{eq:dynamic_interaction_weights} adaptively regulates the feature coupling between the speckle reduction and boundary enhancement branches, which is then propagated to the subsequent encoding stages.

\subsection{Training Objectives}
The NBGL framework is trained through alternating optimization of the generator and the discriminator. The generator is optimized to estimate the despeckled volume $\hat{\mathbf{Y}}$ and the auxiliary boundary probability map $\hat{\mathbf{B}}$, while the patch discriminator is optimized to distinguish real samples from generated samples at the patch level. Through this adversarial training process, the generator is encouraged to generate despeckled outputs with local texture patterns closer to those of the ground-truth volumes.

\subsubsection{Generator Objective}
The generator objective consists of a reconstruction loss $\mathcal{L}_{R}$, a structural similarity loss $\mathcal{L}_{S}$, a boundary preservation loss $\mathcal{L}_{B}$, and an adversarial loss $\mathcal{L}_{A}$. 

\paragraph{Reconstruction Loss}
With $\mathbf{M}$ denoting the foreground mask, the voxel-level image reconstruction term $\mathcal{L}_{R}$ is defined as a masked Charbonnier loss~\citep{lai2017deep}:
\begin{equation}
    \mathcal{L}_{R} = \frac{1}{\sum_i \mathbf{M}_i} 
    \sum_i \mathbf{M}_i 
    \sqrt{\left(\hat{\mathbf{Y}}_i - \mathbf{Y}_i\right)^2 + \epsilon_c^2},
\end{equation}
where $\epsilon_c = 10^{-3}$ is a stability constant for the Charbonnier loss. 

\paragraph{Structural Similarity Loss}
To further preserve structural consistency, a 3D SSIM loss~\citep{wang2004image} is computed within the minimum enclosing 3D bounding box $\Omega(\mathbf{M})$ of the foreground region:
\begin{equation}
    \mathcal{L}_{S} = 1 - \mathrm{SSIM}_{3\mathrm{D}} \big(
    \hat{\mathbf{Y}}_{\Omega(\mathbf{M})},
    \mathbf{Y}_{\Omega(\mathbf{M})}
    \big),
\end{equation}
where $\hat{\mathbf{Y}}_{\Omega(\mathbf{M})}$ and $\mathbf{Y}_{\Omega(\mathbf{M})}$ denote the predicted and GT volumes restricted to $\Omega(\mathbf{M})$, respectively. 

\paragraph{Boundary Preservation Loss}
The boundary preservation loss is evaluated within the anatomical region enclosed by the GT boundary, whose voxel index set is denoted as $\mathcal{I}_{B}$. Within this region, the BCE and soft Dice terms~\citep{milletari2016v} are defined as:
\begin{align}
    \mathcal{L}_{\mathrm{BCE}} &=
    -\frac{1}{|\mathcal{I}_{B}|}
    \sum_{i\in\mathcal{I}_{B}}
    \left[
    B_i \log(\hat{B}_i+\epsilon_b)
    +
    (1-B_i)\log(1-\hat{B}_i+\epsilon_b)
    \right],
    \label{eq:bce_loss} \\
    \mathcal{L}_{\mathrm{Dice}} &=
    1 -
    \frac{
    2\sum_{i\in\mathcal{I}_{B}} \hat{B}_i B_i + \epsilon_d
    }{
    \sum_{i\in\mathcal{I}_{B}} \hat{B}_i
    +
    \sum_{i\in\mathcal{I}_{B}} B_i
    + \epsilon_d
    },
    \label{eq:dice_loss}
\end{align}
where $\epsilon_b$ and $\epsilon_d$ are stability constants. 
In this study, $\epsilon_b=\epsilon_d=10^{-6}$. The resulting boundary preservation loss is defined as:
\begin{equation}
    \mathcal{L}_{\mathrm{B}}
    =
    \mathcal{L}_{\mathrm{BCE}}
    +
    \mathcal{L}_{\mathrm{Dice}}.
    \label{eq:boundary_loss}
\end{equation}

\paragraph{Adversarial Loss}
To capture high-frequency textural patterns and enforce fine-grained structural alignment, we incorporate a conditional 3D patch-level discriminator $\mathcal{D}$ inspired by~\citep{isola2017image}. During generator optimization, $\mathcal{D}$ is fixed, and the generator is optimized to make the synthesized pair $(\mathbf{X}, \hat{\mathbf{Y}})$ be classified as real. Following the least-squares GAN formulation~\citep{mao2017least}, the adversarial loss for the generator is defined as:
\begin{equation}
    \mathcal{L}_{A}
    =
    \frac{1}{2}\mathbb{E}_{\mathbf{X}}
    \big[
    (\mathcal{D}(\mathbf{X}, \hat{\mathbf{Y}}) - 1)^2
    \big].
    \label{eq:adv_loss}
\end{equation}

\paragraph{Overall Generator Loss}
The overall generator loss is: 
\begin{equation}
    \mathcal{L}_{G} = 
    \mathcal{L}_{R}
    + \lambda_{S}\mathcal{L}_{S}
    + \lambda_{B}\mathcal{L}_{B}     
    + \lambda_{A}\mathcal{L}_{A}, 
    \label{eq:overall_loss}
\end{equation}
where $\lambda_{S}$, $\lambda_{B}$, and $\lambda_{A}$ are weighting coefficients for the structural similarity loss, boundary preservation loss, and adversarial loss, respectively. The first two losses constrain the despeckled volume $\hat{\mathbf{Y}}$, $\mathcal{L}_{B}$ supervises the predicted anatomical boundary map $\hat{\mathbf{B}}$, and $\mathcal{L}_{A}$ encourages patch-level texture consistency between the despeckled and GT volumes.

\subsubsection{Discriminator Objective}
The discriminator $\mathcal{D}$ is optimized alternately with the generator to distinguish real pairs $(\mathbf{X}, \mathbf{Y})$ from generated pairs $(\mathbf{X}, \hat{\mathbf{Y}})$. Its objective is defined as:
\begin{equation}
    \mathcal{L}_{D} = \frac{1}{2}\mathbb{E}_{\mathbf{X}, \mathbf{Y}}
    \big[
    (\mathcal{D}(\mathbf{X}, \mathbf{Y}) - 1)^2
    \big]
    +
    \frac{1}{2}\mathbb{E}_{\mathbf{X}}
    \big[
    \mathcal{D}(\mathbf{X}, \hat{\mathbf{Y}})^2
    \big].
    \label{eq:disc_loss}
\end{equation}

\begin{table*}[!t]
    \centering
    \caption{Quantitative comparison of the proposed NBGL against 12 baseline models for ultrasound speckle reduction on the UterUS test set under low noise levels ($0.005$ and $0.01$). The best results are highlighted in bold; $p_{\mathrm{Holm}}$ denotes Holm-adjusted $p$-values.}
    \label{tab:comparison_low_noise}
    \renewcommand{\arraystretch}{0.7}
    \resizebox{\textwidth}{!}{%
    \begin{tabular}{l|cccc|cccc}
        \toprule
        \multirow{2}{*}{Method}
        & \multicolumn{4}{c|}{Noise level = 0.005}
        & \multicolumn{4}{c}{Noise level = 0.01} \\
        \cmidrule(lr){2-5} \cmidrule(lr){6-9}
        & PSNR$\uparrow$ 
        & SSIM$\uparrow$ 
        & RMSE$\downarrow$ 
        & $p_{\mathrm{Holm}}$
        & PSNR$\uparrow$ 
        & SSIM$\uparrow$ 
        & RMSE$\downarrow$ 
        & $p_{\mathrm{Holm}}$ \\
        \midrule
        Gaussian Filter
        & {25.3663}{\footnotesize $\pm$1.8395}  
        & {0.9229}{\footnotesize $\pm$0.0152}  
        & {0.0550}{\footnotesize $\pm$0.0110}  
        & {$8.94\times 10^{-8}$} 
        & {25.1312}{\footnotesize $\pm$1.8498}  
        & {0.9196}{\footnotesize $\pm$0.0159}  
        & {0.0566}{\footnotesize $\pm$0.0114}  
        & {$8.94\times 10^{-8}$}  \\
        Mean Filter
        & {22.4154}{\footnotesize $\pm$1.7618}  
        & {0.8381}{\footnotesize $\pm$0.0290}  
        & {0.0772}{\footnotesize $\pm$0.0148}  
        & {$8.94\times 10^{-8}$} 
        & {22.3585}{\footnotesize $\pm$1.7714}  
        & {0.8368}{\footnotesize $\pm$0.0293}  
        & {0.0777}{\footnotesize $\pm$0.0150}  
        & {$8.94\times 10^{-8}$}  \\
        Median Filter
        & {23.0198}{\footnotesize $\pm$1.6702}  
        & {0.8576}{\footnotesize $\pm$0.0250}  
        & {0.0719}{\footnotesize $\pm$0.0133}  
        & {$8.94\times 10^{-8}$} 
        & {22.8982}{\footnotesize $\pm$1.6632}  
        & {0.8552}{\footnotesize $\pm$0.0250}  
        & {0.0729}{\footnotesize $\pm$0.0134}  
        & {$8.94\times 10^{-8}$}  \\
        Bilateral Filter
        & {28.3858}{\footnotesize $\pm$0.9057}  
        & {0.9567}{\footnotesize $\pm$0.0082}  
        & {0.0383}{\footnotesize $\pm$0.0039}  
        & {$8.94\times 10^{-8}$} 
        & {27.2901}{\footnotesize $\pm$1.2434}  
        & {0.9499}{\footnotesize $\pm$0.0079}  
        & {0.0436}{\footnotesize $\pm$0.0062}  
        & {$8.94\times 10^{-8}$}  \\
        BM4D
        & {29.8252}{\footnotesize $\pm$2.0826}  
        & {0.9797}{\footnotesize $\pm$0.0107}  
        & {0.0332}{\footnotesize $\pm$0.0080}  
        & {2.04$\times 10^{-6}$} 
        & {26.8829}{\footnotesize $\pm$2.0320}  
        & {0.9636}{\footnotesize $\pm$0.0173}  
        & {0.0465}{\footnotesize $\pm$0.0109}  
        & {$8.94\times 10^{-8}$} \\
        \midrule
        DESD-GAN
        & {30.1461}{\footnotesize $\pm$2.3742}  
        & {0.9778}{\footnotesize $\pm$0.0128}  
        & {0.0322}{\footnotesize $\pm$0.0088}  
        & {$8.94\times 10^{-8}$}
        & {28.6554}{\footnotesize $\pm$2.1444}  
        & {0.9690}{\footnotesize $\pm$0.0157}  
        & {0.0380}{\footnotesize $\pm$0.0093}  
        & {$8.94\times 10^{-8}$} \\
        DU-GAN
        & {29.4820}{\footnotesize $\pm$1.4971}  
        & {0.9720}{\footnotesize $\pm$0.0138}  
        & {0.0340}{\footnotesize $\pm$0.0057}  
        & {$2.04\times 10^{-6}$}
        & {27.7330}{\footnotesize $\pm$1.1997}  
        & {0.9614}{\footnotesize $\pm$0.0145}  
        & {0.0414}{\footnotesize $\pm$0.0060}  
        & {$2.46\times 10^{-7}$} \\
        G2CR-FPN
        & {24.4753}{\footnotesize $\pm$1.1896}  
        & {0.7431}{\footnotesize $\pm$0.0459}  
        & {0.0603}{\footnotesize $\pm$0.0082}  
        & {$8.94\times 10^{-8}$} 
        & {24.2674}{\footnotesize $\pm$1.2203}  
        & {0.7397}{\footnotesize $\pm$0.0457}  
        & {0.0618}{\footnotesize $\pm$0.0086}  
        & {$8.94\times 10^{-8}$} \\
        LIT-Former
        & {10.6689}{\footnotesize $\pm$1.9520}  
        & {0.4977}{\footnotesize $\pm$0.0359}  
        & {0.3000}{\footnotesize $\pm$0.0677}  
        & {$8.94\times 10^{-8}$}
        & {10.6761}{\footnotesize $\pm$1.9527}  
        & {0.4981}{\footnotesize $\pm$0.0358}  
        & {0.2998}{\footnotesize $\pm$0.0677}  
        & {$8.94\times 10^{-8}$} \\
        Speckle2Self
        & {17.0142}{\footnotesize $\pm$2.3782}  
        & {0.6493}{\footnotesize $\pm$0.0344}  
        & {0.1464}{\footnotesize $\pm$0.0422}  
        & {$8.94\times 10^{-8}$}
        & {17.0332}{\footnotesize $\pm$2.3629}  
        & {0.6493}{\footnotesize $\pm$0.0344}  
        & {0.1460}{\footnotesize $\pm$0.0419}  
        & {$8.94\times 10^{-8}$} \\
        Autoencoder-KL
        & {19.7913}{\footnotesize $\pm$0.9836}  
        & {0.5761}{\footnotesize $\pm$0.1635}  
        & {0.1031}{\footnotesize $\pm$0.0115}  
        & {$8.94\times 10^{-8}$}
        & {19.7829}{\footnotesize $\pm$0.9901}  
        & {0.5755}{\footnotesize $\pm$0.1638}  
        & {0.1032}{\footnotesize $\pm$0.0116}  
        & {$8.94\times 10^{-8}$} \\
        DDIM
        & {16.0034}{\footnotesize $\pm$3.6688}  
        & {0.6426}{\footnotesize $\pm$0.1480}  
        & {0.1730}{\footnotesize $\pm$0.0760}  
        & {$8.94\times 10^{-8}$}
        & {16.3888}{\footnotesize $\pm$3.6070}  
        & {0.6241}{\footnotesize $\pm$0.1448}  
        & {0.1651}{\footnotesize $\pm$0.0722}  
        & {$8.94\times 10^{-8}$} \\
        \midrule
        \textbf{NBGL (Ours)}
        & \textbf{{31.5758}{\footnotesize $\pm$1.9705}} 
        & \textbf{{0.9856}{\footnotesize $\pm$0.0062}} 
        & \textbf{{0.0270}{\footnotesize $\pm$0.0060}} 
        & \textbf{--}
        & \textbf{{29.6079}{\footnotesize $\pm$1.9295}} 
        & \textbf{{0.9775}{\footnotesize $\pm$0.0093}} 
        & \textbf{{0.0339}{\footnotesize $\pm$0.0074}} 
        & \textbf{--} \\
        \bottomrule
    \end{tabular}
    }
\end{table*}

\subsection{Implementation Details}
\label{sec:implementation}
The proposed NBGL framework was implemented in PyTorch and trained for 200 epochs on NVIDIA A100 GPUs with automatic mixed precision. 
The generator and discriminator were optimized alternately using Adam with an initial learning rate of $10^{-4}$.
Validation was performed after each epoch, and the best-performing model parameters on the validation set were used for final evaluation. To stabilize early training, we used an epoch-wise warm-up strategy for the adversarial term. Specifically, the adversarial loss weight $\lambda_A$ was set to 0 for the first 10 epochs, increased to 0.01 until epoch 30, and fixed at 0.02 thereafter. This schedule allows the reconstruction and boundary objectives to provide stable supervision before stronger adversarial regularization is introduced.
The loss weights in Eq.~\eqref{eq:overall_loss} were set to $\lambda_S=4.0$ and $\lambda_B=0.5$.  
Parameter sensitivity analysis is detailed in Section~\ref{sec:loss_weight}.
For the noise-aware interaction weight generation module, the estimated speckle noise level was normalized to $[0,1]$ using lower and upper bounds of 0.003 and 0.12, respectively. 
The interaction weight was bounded by $\omega_{\min}=0.5$ and $\omega_{\max}=1.5$, and the power-law exponent was set to $\rho=2.0$. The direction-specific scaling factors were set to $\alpha^{(2)}=1.0$ and $\alpha^{(3)}=0.2$.

During \emph{training}, we randomly sampled $192\times192\times96$ patches from each volume, including one boundary-centered patch and additional random patches, with the number of patches determined by the relative sizes of the volume and the patch. 
For adversarial supervision, $64\times64\times64$ sub-patches were sampled from the foreground region and used as inputs to the 3D patch-level discriminator. 
During \emph{inference}, the trained model takes only the speckled volume as input, without requiring the annotation mask or GT boundary map. 
Full-volume inference was performed using 3D sliding-window inference. 
Each window covered 64 voxels along the sliding direction, with adjacent windows overlapping by 16 voxels. 
The overlapping predictions were combined using Hann-window blending. 
To reduce directional bias introduced by sliding-window inference, predictions were independently obtained by sliding along each of the three spatial dimensions and then averaged to obtain the final despeckled volume.

\section{Experiments}
\label{sec:experiments}
\subsection{Speckle Noise Simulation}
Ultrasound speckle is commonly modeled as multiplicative noise~\citep{sivaanpu2025speckle}. To simulate paired speckled and GT volumes, synthetic speckle is introduced into the GT volumes. 
Let $y$ denote the intensity of a foreground voxel in the GT volume, and let $x$ denote the corresponding speckled intensity. 
In the logarithmic domain, the multiplicative speckle model can be written as:
\begin{equation}
    \log(x) = \log(y) + \log(\eta),
\end{equation}
where $\eta$ denotes the multiplicative speckle term. 
However, if $\log(\eta)$ is sampled directly from $\mathcal{N}(0,\sigma^2)$, $\eta$ follows a log-normal distribution with 
$\mathbb{E}[\eta]=\exp(\sigma^2/2)>1$, which introduces an intensity bias after exponentiation and shifts the global intensity distribution of the simulated volumes~\citep{krishnamoorthy2003inferences}. 
This bias can be corrected by sampling the log-speckle term from a Gaussian distribution with mean correction to ensure that $\mathbb{E}[\eta]=1$:
\begin{equation}
    \log(x) = \log(y) + \log(\eta),
    \quad 
    \log(\eta) \sim \mathcal{N}\left(-\frac{\sigma^2}{2}, \sigma^2\right),
\label{eq:mean_preserving_noise_simulation}
\end{equation}
where $\sigma^2$ denotes the log-speckle noise variance.

\subsection{Experimental Settings}
\paragraph{Dataset and Preprocessing}
Experiments were conducted on the UterUS dataset~\citep{bonevs2024automatic}, comprising 141 3D transvaginal ultrasound volumes with expert annotations of the uterine cavity, partitioned into 85 training, 28 validation, and 28 test volumes. Detailed subject IDs for all data partitions are listed in the \emph{Supplementary Materials} for reproducibility. Data preprocessing involved extracting a foreground mask $\mathbf{M}$ from the GT volume using an intensity threshold of $10^{-6}$. Voxels within $\mathbf{M}$ were min-max normalized to $[0, 1]$, and voxels outside $\mathbf{M}$ were set to zero. 
The GT boundary map $\mathbf{B}$ used for boundary supervision was extracted from the binary annotation mask $\mathbf{S}$ by subtracting its eroded version. 
The erosion was implemented with a fixed $3\times3\times3$ cubic structuring element to extract the inner boundary of the annotation mask. 
Synthetic speckle was generated using the mean-preserving speckle simulation model in Eq.~\eqref{eq:mean_preserving_noise_simulation} with six variance levels, $\sigma^2 \in \{0.005, 0.01, 0.02, 0.05, 0.1, 0.2\}$, and was applied only within the foreground mask $\mathbf{M}$. 
During training, all six noise levels were applied to each training volume to construct a mixed-noise training set, and the resulting speckled-GT pairs were combined to train the speckle reduction model.  
During inference, each test volume was evaluated separately at each noise level, and the results were also aggregated across all six levels for overall evaluation.

\begin{table*}[!t]
\setlength{\abovecaptionskip}{0pt}
\setlength{\belowcaptionskip}{0pt}
\setlength{\abovedisplayskip}{0pt}
\setlength{\belowdisplayskip}{0pt}
    \centering
    \caption{Quantitative comparison of the proposed NBGL against 12 baseline models for ultrasound speckle reduction on the UterUS test set under moderate noise levels ($0.02$ and $0.05$). The best results are highlighted in bold; $p_{\mathrm{Holm}}$ denotes Holm-adjusted $p$-values.}
     \renewcommand{\arraystretch}{0.7}   
     \label{tab:comparison_intermediate_noise}
    \resizebox{\textwidth}{!}{%
    \begin{tabular}{l|cccc|cccc}
        \toprule
        \multirow{2}{*}{Method}
        & \multicolumn{4}{c|}{Noise level = 0.02}
        & \multicolumn{4}{c}{Noise level = 0.05} \\
        \cmidrule(lr){2-5} \cmidrule(lr){6-9}
        & PSNR$\uparrow$ 
        & SSIM$\uparrow$ 
        & RMSE$\downarrow$ 
        & $p_{\mathrm{Holm}}$
        & PSNR$\uparrow$ 
        & SSIM$\uparrow$ 
        & RMSE$\downarrow$ 
        & $p_{\mathrm{Holm}}$ \\
        \midrule
        Gaussian Filter
        & {24.7092}{\footnotesize $\pm$1.8719}  
        & {0.9137}{\footnotesize $\pm$0.0175} 
        & {0.0594}{\footnotesize $\pm$0.0121}  
        & {$8.94\times 10^{-8}$}
        & {23.6967}{\footnotesize $\pm$1.9302}    
        & {0.8992}{\footnotesize $\pm$0.0218}  
        & {0.0669}{\footnotesize $\pm$0.0142}  
        & {$8.94\times 10^{-8}$} \\
        Mean Filter
        & {22.2483}{\footnotesize $\pm$1.7915}  
        & {0.8346}{\footnotesize $\pm$0.0298} 
        & {0.0787}{\footnotesize $\pm$0.0153}  
        & {$8.94\times 10^{-8}$}
        & {21.9348}{\footnotesize $\pm$1.8509}    
        & {0.8288}{\footnotesize $\pm$0.0309}  
        & {0.0817}{\footnotesize $\pm$0.0164}  
        & {$8.94\times 10^{-8}$}\\
        Median Filter
        & {22.6752}{\footnotesize $\pm$1.6571}  
        & {0.8512}{\footnotesize $\pm$0.0250} 
        & {0.0748}{\footnotesize $\pm$0.0136}  
        & {$8.94\times 10^{-8}$}
        & {22.0867}{\footnotesize $\pm$1.6622}    
        & {0.8415}{\footnotesize $\pm$0.0254}  
        & {0.0800}{\footnotesize $\pm$0.0145}  
        & {$8.94\times 10^{-8}$} \\
        Bilateral Filter
        & {25.3897}{\footnotesize $\pm$1.6383}  
        & {0.9343}{\footnotesize $\pm$0.0129} 
        & {0.0547}{\footnotesize $\pm$0.0102}  
        & {$8.94\times 10^{-8}$}
        & {21.7550}{\footnotesize $\pm$1.8932}    
        & {0.8918}{\footnotesize $\pm$0.0268}  
        & {0.0836}{\footnotesize $\pm$0.0180}  
        & {$8.94\times 10^{-8}$} \\
        BM4D
        & {23.9852}{\footnotesize $\pm$1.9571}  
        & {0.9379}{\footnotesize $\pm$0.0257} 
        & {0.0648}{\footnotesize $\pm$0.0146}  
        & {$8.94\times 10^{-8}$}
        & {20.2691}{\footnotesize $\pm$1.8181}    
        & {0.8863}{\footnotesize $\pm$0.0366}  
        & {0.0990}{\footnotesize $\pm$0.0206}  
        & {$8.94\times 10^{-8}$} \\
        \midrule
        DESD-GAN
        & {27.1225}{\footnotesize $\pm$1.9840}  
        & {0.9557}{\footnotesize $\pm$0.0197} 
        & {0.0451}{\footnotesize $\pm$0.0101}  
        & {$8.94\times 10^{-8}$}
        & {25.2265}{\footnotesize $\pm$1.8378}    
        & {0.9305}{\footnotesize $\pm$0.0263}  
        & {0.0559}{\footnotesize $\pm$0.0114}  
        & {$8.94\times 10^{-8}$} \\
        DU-GAN
        & {26.0707}{\footnotesize $\pm$1.2195}  
        & {0.9455}{\footnotesize $\pm$0.0171} 
        & {0.0502}{\footnotesize $\pm$0.0074}  
        & {$8.94\times 10^{-8}$}
        & {24.1750}{\footnotesize $\pm$1.3532}    
        & {0.9162}{\footnotesize $\pm$0.0221}  
        & {0.0626}{\footnotesize $\pm$0.0098}  
        & {$8.94\times 10^{-8}$} \\
        G2CR-FPN
        & {23.9368}{\footnotesize $\pm$1.2607}  
        & {0.7343}{\footnotesize $\pm$0.0453} 
        & {0.0642}{\footnotesize $\pm$0.0092}  
        & {$8.94\times 10^{-8}$}
        & {23.2802}{\footnotesize $\pm$1.3221}    
        & {0.7228}{\footnotesize $\pm$0.0445}  
        & {0.0693}{\footnotesize $\pm$0.0103}  
        & {$8.94\times 10^{-8}$} \\
        LIT-Former
        & {10.6881}{\footnotesize $\pm$1.9556}  
        & {0.4988}{\footnotesize $\pm$0.0358} 
        & {0.2994}{\footnotesize $\pm$0.0677}  
        & {$8.94\times 10^{-8}$}
        & {10.7088}{\footnotesize $\pm$1.9576}    
        & {0.5000}{\footnotesize $\pm$0.0357}  
        & {0.2987}{\footnotesize $\pm$0.0677}  
        & {$8.94\times 10^{-8}$} \\
        Speckle2Self
        & {17.0601}{\footnotesize $\pm$2.3414}  
        & {0.6493}{\footnotesize $\pm$0.0344} 
        & {0.1454}{\footnotesize $\pm$0.0413}  
        & {$8.94\times 10^{-8}$}
        & {17.1019}{\footnotesize $\pm$2.2965}    
        & {0.6492}{\footnotesize $\pm$0.0345}  
        & {0.1446}{\footnotesize $\pm$0.0403}  
        & {$8.94\times 10^{-8}$} \\
        Autoencoder-KL
        & {19.7552}{\footnotesize $\pm$0.9884}  
        & {0.5753}{\footnotesize $\pm$0.1641} 
        & {0.1035}{\footnotesize $\pm$0.0116}  
        & {$8.94\times 10^{-8}$}
        & {19.6451}{\footnotesize $\pm$0.9867}    
        & {0.5743}{\footnotesize $\pm$0.1640}  
        & {0.1048}{\footnotesize $\pm$0.0118}  
        & {$8.94\times 10^{-8}$} \\
        DDIM
        & {16.0098}{\footnotesize $\pm$3.7958}  
        & {0.6124}{\footnotesize $\pm$0.1298} 
        & {0.1737}{\footnotesize $\pm$0.0772}  
        & {$8.94\times 10^{-8}$}
        & {15.5916}{\footnotesize $\pm$3.3714}    
        & {0.5874}{\footnotesize $\pm$0.1271}  
        & {0.1789}{\footnotesize $\pm$0.0721}  
        & {$8.94\times 10^{-8}$} \\
        \midrule
        \textbf{NBGL (Ours)}
        & \textbf{{27.8091}{\footnotesize $\pm$1.9000}} 
        & \textbf{{0.9658}{\footnotesize $\pm$0.0132}} 
        & \textbf{{0.0416}{\footnotesize $\pm$0.0089}} 
        & \textbf{--}
        & \textbf{{25.7602}{\footnotesize $\pm$1.8442}}  
        & \textbf{{0.9447}{\footnotesize $\pm$0.0191}} 
        &\textbf{{0.0526}{\footnotesize $\pm$0.0108}} 
        & \textbf{--} \\
        \bottomrule
    \end{tabular}
    }
\end{table*}

\paragraph{Competing Methods}
The NBGL is compared against twelve competing methods, comprising five filtering-based approaches and seven deep learning-based approaches. 
The filtering methods comprise four local spatial-domain filters, including \textbf{Gaussian Filter}~\citep{jahne2005digital}, \textbf{Mean Filter}~\citep{jahne2005digital}, \textbf{Median Filter}~\citep{huang1979fast}, and \textbf{Bilateral Filter}~\citep{tomasi1998bilateral}, as well as one non-local transform-domain method \textbf{BM4D}~\citep{maggioni2012nonlocal}. 
The deep learning-based methods include representative models from image denoising and restoration, ultrasound speckle reduction, and generative image modeling, including \textbf{DESD-GAN}~\citep{han2022dual}, 
\textbf{DU-GAN}~\citep{huang2021gan}, 
\textbf{G2CR-FPN}~\citep{zhou2024gradient}, 
\textbf{LIT-Former}~\citep{chen2024lit}, \textbf{Speckle2Self}~\citep{li2025speckle2self}, 
\textbf{Autoencoder-KL}~\citep{rombach2022high}, and \textbf{DDIM}~\citep{song2021denoising}. 

All filtering-based methods were directly applied to the 3D ultrasound volumes. Their hyperparameters were selected by grid search on the validation set. For BM4D, the noise standard deviation parameter was selected by grid search on the validation set, while other BM4D parameters were kept at default settings. For deep learning-based baselines originally developed for 2D images, including DESD-GAN, DU-GAN, G2CR-FPN, and Speckle2Self, 2D axial slices were used for training and validation, as this reduced padding for the fixed patch size. During testing, these methods were applied slice-wise along the three orthogonal axes of each volume, and the final 3D outputs were obtained by averaging the three axis-wise predictions before quantitative evaluation. Methods compatible with volumetric inputs were trained directly on 3D patches. For the self-supervised despeckling baseline Speckle2Self, the original network frameworks were retained, but their training objectives were adapted to the same paired supervised setting as other learning-based baselines. For the diffusion baseline DDIM, the speckled volume was used as the conditioning input.

For a fair comparison, all competing methods, including both 2D and 3D methods, used the same training, validation, and test splits and the same six noise-level settings. During training and inference, a consistent patch sampling strategy was adopted across methods. During inference, patch-level predictions were assembled into full volumes using the same reconstruction strategy, and all quantitative metrics were computed on the reconstructed volumes.
Implementation details of the comparison methods are provided in  \emph{Supplementary Materials}.  

\paragraph{Evaluation Metrics}
Quantitative performance was assessed using peak signal-to-noise ratio (PSNR), 3D structural similarity index measure (SSIM), and root mean square error (RMSE). 
PSNR and RMSE were computed within the foreground mask $\mathbf{M}$, while SSIM was computed within the minimum enclosing 3D bounding box $\Omega(\mathbf{M})$ of the foreground region. 
Statistical significance was assessed using two-sided Wilcoxon signed-rank tests on paired SSIM scores between NBGL and each competing method. 
For each individual noise level, the test was performed across the test volumes. 
For the aggregated results across multiple noise levels, SSIM scores were first averaged over all noise levels for each test volume and each method, and the Wilcoxon signed-rank tests were then performed on the resulting paired volume-level averages. To account for multiple pairwise comparisons, $p$-values were adjusted using the Holm-Bonferroni correction, and adjusted $p_{\mathrm{Holm}}<0.05$ was considered statistically significant.

\subsection{Quantitative Results}
Detailed quantitative comparisons under different noise levels ($\sigma^2 \in \{0.005, 0.01, 0.02, 0.05, 0.1, 0.2\}$) are provided in Tables~\ref{tab:comparison_low_noise}--\ref{tab:comparison_high_noise},
while the aggregated results across all six levels are summarized in Table~\ref{tab:comparison_mixed}. 
These multi-level evaluations confirm that our NBGL achieves the top-tier overall performance, consistently yielding the highest PSNR and SSIM alongside the lowest RMSE. 
In particular, the SSIM improvements achieved by NBGL over all competing methods are statistically significant ($p_{\mathrm{Holm}} < 0.05$ after Holm-Bonferroni correction).

\paragraph{Performance at Extreme Noise Levels} 
To demonstrate operational stability, we analyze performance at both ends of the evaluated noise spectrum. 
At the lowest and highest noise levels, NBGL outperforms the strongest filtering-based and deep learning-based baselines. 
At the \emph{lowest noise level} ($\sigma^2$=$0.005$), BM4D and DESD-GAN achieve the best performance among the filtering-based and learning-based methods, respectively. 
NBGL outperforms both, exceeding BM4D by 1.7506~dB in PSNR and 0.0059 in SSIM (with a 0.0062 reduction in RMSE), and surpassing DESD-GAN by 1.4297~dB in PSNR and 0.0078 in SSIM. 
Under \emph{severe noise conditions} ($\sigma^2$=$0.2$) where BM4D degrades drastically, the Gaussian filter becomes the traditional alternative.
And NBGL has prominent gains over the Gaussian filter (+2.3518~dB PSNR, +0.0476 SSIM, and -0.0230 RMSE). 
When compared to the robust deep learning method (DESD-GAN), NBGL further improves PSNR by 0.2581 and SSIM by 0.0215, while reducing RMSE by 0.0020. 
These findings verify that NBGL maintains a consistent performance advantage across the entire evaluated noise range.

\begin{table*}[!h]
    \centering
    \renewcommand{\arraystretch}{0.7}    
    \caption{Quantitative comparison of the proposed NBGL against 12 baseline models for ultrasound speckle reduction on the UterUS test set under severe noise levels ($0.1$ and $0.2$). The best results are highlighted in bold; $p_{\mathrm{Holm}}$ denotes Holm-adjusted $p$-values.}
    \label{tab:comparison_high_noise}
    \resizebox{\textwidth}{!}{%
    \begin{tabular}{l|cccc|cccc}
        \toprule  
        \multirow{2}{*}{Method}
        & \multicolumn{4}{c|}{Noise level = 0.1}
        & \multicolumn{4}{c}{Noise level = 0.2} \\
        \cmidrule(lr){2-5} \cmidrule(lr){6-9}
        & PSNR$\uparrow$ 
        & SSIM$\uparrow$ 
        & RMSE$\downarrow$ 
        & $p_{\mathrm{Holm}}$
        & PSNR$\uparrow$ 
        & SSIM$\uparrow$ 
        & RMSE$\downarrow$ 
        & $p_{\mathrm{Holm}}$ \\
        \midrule
        Gaussian Filter
        & {22.4710}{\footnotesize $\pm$2.0122}  
        & {0.8803}{\footnotesize $\pm$0.0268} 
        & {0.0772}{\footnotesize $\pm$0.0174}  
        & {$8.94\times 10^{-8}$}
        & {20.8298}{\footnotesize $\pm$2.1397}    
        & {0.8531}{\footnotesize $\pm$0.0321}  
        & {0.0936}{\footnotesize $\pm$0.0229}  
        & {$8.94\times 10^{-8}$} \\
        Mean Filter
        & {21.4469}{\footnotesize $\pm$1.9572}  
        & {0.8209}{\footnotesize $\pm$0.0325} 
        & {0.0867}{\footnotesize $\pm$0.0186}  
        & {$8.94\times 10^{-8}$}
        & {20.5831}{\footnotesize $\pm$2.1594}    
        & {0.8081}{\footnotesize $\pm$0.0348}  
        & {0.0963}{\footnotesize $\pm$0.0233}  
        & {$8.94\times 10^{-8}$} \\
        Median Filter
        & {21.2200}{\footnotesize $\pm$1.7015}  
        & {0.8284}{\footnotesize $\pm$0.0265} 
        & {0.0885}{\footnotesize $\pm$0.0164}  
        & {$8.94\times 10^{-8}$}
        & {19.7508}{\footnotesize $\pm$1.7990}    
        & {0.8072}{\footnotesize $\pm$0.0287}  
        & {0.1050}{\footnotesize $\pm$0.0208}  
        & {$8.94\times 10^{-8}$} \\
        Bilateral Filter
        & {18.7165}{\footnotesize $\pm$1.8301}  
        & {0.8426}{\footnotesize $\pm$0.0353} 
        & {0.1184}{\footnotesize $\pm$0.0247}  
        & {$8.94\times 10^{-8}$}
        & {15.8763}{\footnotesize $\pm$1.6897}    
        & {0.7853}{\footnotesize $\pm$0.0380}  
        & {0.1637}{\footnotesize $\pm$0.0315}  
        & {$8.94\times 10^{-8}$} \\
        BM4D
        & {17.6102}{\footnotesize $\pm$1.6873}  
        & {0.8358}{\footnotesize $\pm$0.0414} 
        & {0.1341}{\footnotesize $\pm$0.0258}  
        & {$8.94\times 10^{-8}$}
        & {15.1582}{\footnotesize $\pm$1.5657}    
        & {0.7807}{\footnotesize $\pm$0.0414}  
        & {0.1774}{\footnotesize $\pm$0.0317}  
        & {$8.94\times 10^{-8}$} \\
        \midrule
        DESD-GAN
        & {23.9967}{\footnotesize $\pm$1.7275}  
        & {0.9070}{\footnotesize $\pm$0.0304} 
        & {0.0643}{\footnotesize $\pm$0.0122}  
        & {$8.94\times 10^{-8}$}
        & {22.9235}{\footnotesize $\pm$1.6484}    
        & {0.8792}{\footnotesize $\pm$0.0342}  
        & {0.0726}{\footnotesize $\pm$0.0131}  
        & {$8.94\times 10^{-8}$} \\
        DU-GAN
        & {22.9159}{\footnotesize $\pm$1.4306}  
        & {0.8892}{\footnotesize $\pm$0.0256} 
        & {0.0724}{\footnotesize $\pm$0.0117}  
        & {$8.94\times 10^{-8}$}
        & {21.6646}{\footnotesize $\pm$1.5519}    
        & {0.8612}{\footnotesize $\pm$0.0286}  
        & {0.0838}{\footnotesize $\pm$0.0146}  
        & {$8.94\times 10^{-8}$} \\
        G2CR-FPN
        & {22.6117}{\footnotesize $\pm$1.3739}  
        & {0.7111}{\footnotesize $\pm$0.0438} 
        & {0.0749}{\footnotesize $\pm$0.0115}  
        & {$8.94\times 10^{-8}$}
        & {21.6989}{\footnotesize $\pm$1.4912}    
        & {0.6962}{\footnotesize $\pm$0.0427}  
        & {0.0834}{\footnotesize $\pm$0.0140}  
        & {$8.94\times 10^{-8}$} \\
        LIT-Former
        & {10.7277}{\footnotesize $\pm$1.9613}  
        & {0.5009}{\footnotesize $\pm$0.0357} 
        & {0.2981}{\footnotesize $\pm$0.0677}  
        & {$8.94\times 10^{-8}$}
        & {10.7509}{\footnotesize $\pm$1.9679}    
        & {0.5020}{\footnotesize $\pm$0.0356}  
        & {0.2973}{\footnotesize $\pm$0.0678}  
        & {$8.94\times 10^{-8}$} \\
        Speckle2Self
        & {17.1190}{\footnotesize $\pm$2.2458}  
        & {0.6491}{\footnotesize $\pm$0.0346} 
        & {0.1441}{\footnotesize $\pm$0.0394}  
        & {$8.94\times 10^{-8}$}
        & {17.0896}{\footnotesize $\pm$2.1718}    
        & {0.6486}{\footnotesize $\pm$0.0347}  
        & {0.1442}{\footnotesize $\pm$0.0382}  
        & {$8.94\times 10^{-8}$} \\
        Autoencoder-KL
        & {19.4757}{\footnotesize $\pm$1.0033}  
        & {0.5718}{\footnotesize $\pm$0.1628} 
        & {0.1069}{\footnotesize $\pm$0.0123}  
        & {$8.94\times 10^{-8}$}
        & {19.1917}{\footnotesize $\pm$1.1170}    
        & {0.5643}{\footnotesize $\pm$0.1593}  
        & {0.1106}{\footnotesize $\pm$0.0144}  
        & {$8.94\times 10^{-8}$} \\
        DDIM
        & {15.1608}{\footnotesize $\pm$3.3390}  
        & {0.5560}{\footnotesize $\pm$0.1275} 
        & {0.1878}{\footnotesize $\pm$0.0755}  
        & {$8.94\times 10^{-8}$}
        & {14.7597}{\footnotesize $\pm$3.3102}    
        & {0.5237}{\footnotesize $\pm$0.1280}  
        & {0.1963}{\footnotesize $\pm$0.0776}  
        & {$8.94\times 10^{-8}$} \\
        \midrule
        \textbf{NBGL (Ours)}
        & \textbf{{24.4174}{\footnotesize $\pm$1.7772}} 
        & \textbf{{0.9244}{\footnotesize $\pm$0.0237}} 
        & \textbf{{0.0613}{\footnotesize $\pm$0.0121}} 
        & \textbf{--}
        & \textbf{{23.1816}{\footnotesize $\pm$1.7355}}  
        & \textbf{{0.9007}{\footnotesize $\pm$0.0280}} 
        &\textbf{{0.0706}{\footnotesize $\pm$0.0136  }} 
        & \textbf{--} \\
        \bottomrule
    \end{tabular}
    }
\end{table*}

\paragraph{Performance Trajectories and Scalability} 
In addition to the level-wise analysis above, performance changes from low to high noise levels further demonstrate the robustness advantage of NBGL under increasing speckle noise. As $\sigma^2$ increases from $0.005$ to $0.2$, BM4D drops by 14.6670 in PSNR and 0.1990 in SSIM, while DESD-GAN drops by 7.2226 in PSNR and 0.0986 in SSIM. Over the same range, NBGL drops by 8.3942 in PSNR and 0.0849 in SSIM, showing a smaller SSIM reduction than the strongest learning-based baseline DESD-GAN and the strongest low-noise filtering baseline BM4D. Although NBGL shows a larger PSNR decrease than DESD-GAN due to its higher initial performance at low noise, its smaller SSIM reduction suggests stronger structural stability under increasing speckle noise. Accordingly, the SSIM difference between NBGL and DESD-GAN increases from 0.0078 at $\sigma^2$=$0.005$ to 0.0215 at $\sigma^2$=$0.2$, indicating that the performance advantage of NBGL over DESD-GAN grows as the noise level increases. This pattern is consistent with the noise-aware mechanism, in which cross-branch coupling is strengthened under a higher estimated noise level to provide more boundary-sensitive guidance to the speckle reduction branch.

\paragraph{Comparison with Structure-Prior Methods} 
Comparisons with baselines that also incorporate structural information, particularly at $\sigma^2$=$0.2$, further support the design of NBGL. G2CR-FPN relies on gradient-based features for structure preservation, but its SSIM drops to 0.6962 under severe speckle noise. NBGL outperforms G2CR-FPN by 1.4827 in PSNR and 0.2045 in SSIM, with a 0.0128 reduction in RMSE. This gap suggests that gradient responses may be strongly distorted by severe speckle noise, limiting their effectiveness for boundary preservation. DU-GAN incorporates edge-aware information through dual-domain discriminators and reaches 0.8612 in SSIM at $\sigma^2$=$0.2$. In comparison, NBGL improves PSNR by 1.5170 and SSIM by 0.0395, while reducing RMSE by 0.0132. A possible explanation is that the wFiLM module incorporates boundary-sensitive features into the speckle reduction branch at intermediate encoder stages, whereas DU-GAN constrains edge-aware features mainly through discriminator-based supervision.

\begin{table}[!t]
    \centering
    \renewcommand{\arraystretch}{0.7}   
    \caption{Quantitative comparison of the proposed NBGL against 12 baseline models for ultrasound speckle reduction on the UterUS test set, aggregated over all six noise levels. The best results are highlighted in bold; $p_{\mathrm{Holm}}$ denotes Holm-adjusted $p$-values.}
    \label{tab:comparison_mixed}
    \setlength{\tabcolsep}{3pt}
\footnotesize  
    \resizebox{\linewidth}{!}{%
    \begin{tabular}{l|ccc|c}
        \toprule
        Method 
        & PSNR$\uparrow$ 
        & SSIM$\uparrow$ 
        & RMSE$\downarrow$ 
        & $p_{\mathrm{Holm}}$ \\
        \midrule
        Gaussian Filter
        & {23.7007}{$\pm$1.9015}
        & {0.8981}{$\pm$0.0206}
        & {0.0681}{$\pm$0.0145}
        & {$8.94\times 10^{-8}$} \\
        Mean Filter
        & {21.8312}{$\pm$1.8604}  
        & {0.8279}{$\pm$0.0307}
        & {0.0831}{$\pm$0.0169}
        & {$8.94\times 10^{-8}$} \\
        Median Filter  
        & {21.9418}{$\pm$1.6617}  
        & {0.8402}{$\pm$0.0254}
        & {0.0822}{$\pm$0.0149}
        & {$8.94\times 10^{-8}$} \\
        Bilateral Filter          
        & {22.9022}{$\pm$1.5153}  
        & {0.8934}{$\pm$0.0189}
        & {0.0837}{$\pm$0.0156}
        & {$8.94\times 10^{-8}$} \\
        BM4D         
        & {22.2885}{$\pm$1.8561}  
        & {0.8973}{$\pm$0.0286}
        & {0.0925}{$\pm$0.0186}
        & {$8.94\times 10^{-8}$} \\
        \midrule
        DESD-GAN  
        & {26.3451}{$\pm$1.9442}  
        & {0.9365}{$\pm$0.0230}
        & {0.0514}{$\pm$0.0107}
        & {$8.94\times 10^{-8}$} \\
        DU-GAN    
        & {25.3402}{$\pm$1.1494}  
        & {0.9243}{$\pm$0.0177}
        & {0.0574}{$\pm$0.0082}
        & {$8.94\times 10^{-8}$} \\
        G2CR-FPN 
        & {23.3784}{$\pm$1.2989}  
        & {0.7245}{$\pm$0.0442}
        & {0.0690}{$\pm$0.0102}
        & {$8.94\times 10^{-8}$} \\
        LIT-Former   
        & {10.7034}{$\pm$1.9578}  
        & {0.4996}{$\pm$0.0357}
        & {0.2989}{$\pm$0.0677}
        & {$8.94\times 10^{-8}$} \\
        Speckle2Self
        & {17.0696}{$\pm$2.2985}  
        & {0.6491}{$\pm$0.0345}
        & {0.1451}{$\pm$0.0406}
        & {$8.94\times 10^{-8}$} \\
        Autoencoder-KL  
        & {19.6070}{$\pm$1.0081}  
        & {0.5729}{$\pm$0.1629}
        & {0.1053}{$\pm$0.0121}
        & {$8.94\times 10^{-8}$} \\
        DDIM
        & {15.6524}{$\pm$3.4621}  
        & {0.5910}{$\pm$0.1290}
        & {0.1791}{$\pm$0.0744}
        & {$8.94\times 10^{-8}$} \\
        \midrule
        \textbf{NBGL (Ours)} 
        & \textbf{{27.0587}{$\pm$1.8528}} 
        & \textbf{{0.9498}{$\pm$0.0165}} 
        & \textbf{{0.0479}{$\pm$0.0098}} 
        & \textbf{--} \\
        \bottomrule
    \end{tabular}
    } 
\end{table}

\paragraph{Aggregated Statistical Analysis}
To evaluate the macro-level stability of our framework, Table~\ref{tab:comparison_mixed}  summarizes the quantitative results aggregated across all six noise levels, capturing the cross-volume mean and standard deviation for each competitor. 
This table confirms that NBGL establishes a new performance ceiling across all metrics, achieving the highest mean PSNR ($27.0587$), the highest mean SSIM ($0.9498$), and the lowest mean RMSE ($0.0479$). When benchmarks are grouped, NBGL yields substantial gains over the top-performing traditional baseline (Gaussian filter), elevating the mean PSNR by $3.3580$~dB and the mean SSIM by $0.0517$. 
More importantly, it consistently surpasses the leading deep learning benchmark (DESD-GAN) by $0.7136$~dB in mean PSNR and $0.0133$ in mean SSIM, confirming its superiority under mixed multi-scale noise conditions. In addition, NBGL shows low SSIM variability, with a standard deviation of 0.0165 in the aggregated results, which is lower than those of both DESD-GAN (0.0230) and the Gaussian filter (0.0206). Although the Gaussian filter also shows relatively small SSIM variability, its mean SSIM is substantially lower than that of NBGL (0.8981 vs. 0.9498), indicating that its stability is achieved at a lower structural similarity level. 
NBGL maintains a high mean SSIM with low variability across the evaluated noise range, suggesting more robust structural preservation under heterogeneous noise levels. This may be attributed to the noise-adaptive cross-branch coupling that adjusts the interaction strength according to the estimated noise level. Two-sided Wilcoxon signed-rank tests with Holm-Bonferroni correction, performed on volume-level SSIM scores averaged over the six noise levels, further confirm that NBGL achieves statistically significant improvements over baselines. 

\begin{figure*}
\setlength{\abovecaptionskip}{0pt}
\setlength{\belowcaptionskip}{0pt}
\setlength{\abovedisplayskip}{0pt}
\setlength{\belowdisplayskip}{0pt}
    \centering
    \includegraphics[width=1\textwidth]{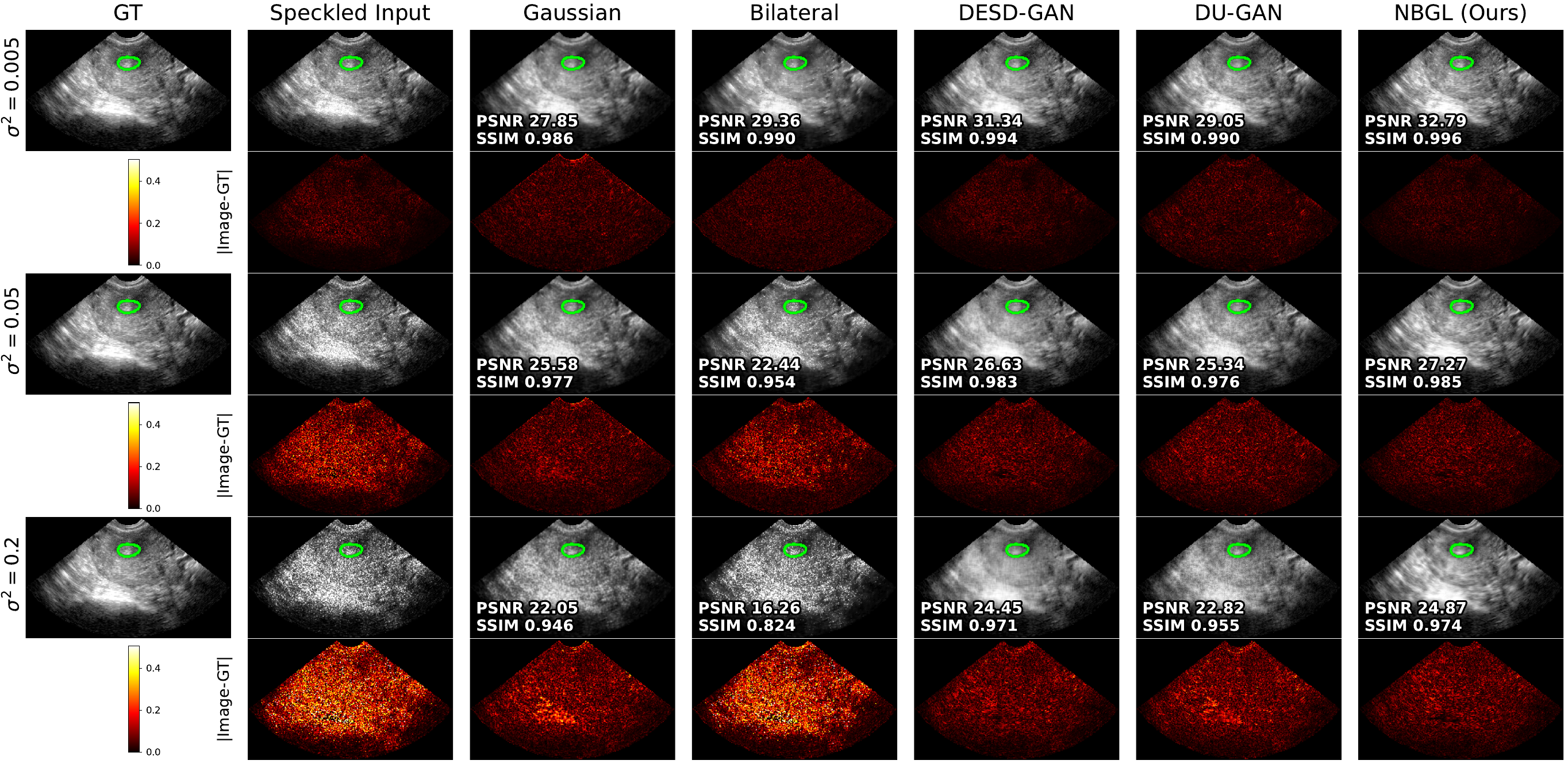}
    \caption{Qualitative comparison of speckle reduction results on a representative sagittal slice under light, moderate, and severe noise levels. The ground-truth (GT) annotated boundary (green line) is overlaid on the speckled input and all despeckled outputs for visual reference. 
    For each noise level, the upper row shows the despeckled outputs and the lower row shows the corresponding absolute error maps with respect to the GT image. The PSNR and SSIM results are reported on each despeckled image for slice-level reference. }    
    \label{fig:qualitative_comparison_error_maps}
\end{figure*}

\subsection{Qualitative Comparison}
Figure~\ref{fig:qualitative_comparison_error_maps} presents the despeckled outputs and absolute error maps of NBGL and the four competing methods 
(\ie, Gaussian filter, Bilateral filter, DESD-GAN, and DU-GAN) on a representative sagittal slice under light ($\sigma^2$=$0.005$), moderate ($\sigma^2$=$0.05$), and severe ($\sigma^2$=$0.2$) noise levels. 
Additional coronal and axial views of the same test volume are provided in \emph{Supplementary Materials}. The ground truth (GT) annotated boundary, corresponding to the uterine cavity boundary, is overlaid on the speckled input and all despeckled outputs to provide a visual reference for the annotated region. A unified color scale is used for all error maps across methods and noise levels, with brighter colors indicating larger deviations from GT. 

At $\sigma^2$=$0.005$, all compared methods suppress most visible speckle, and the differences among learning-based methods are relatively small. Nevertheless, the NBGL shows lower residual errors on the representative slice shown in Fig.~\ref{fig:qualitative_comparison_error_maps}, achieving a slice-level PSNR/SSIM of 32.79/0.996, compared with 31.34/0.994 for the strongest competing baseline, DESD-GAN. As the noise level increases to $\sigma^2$=$0.05$, the differences in the error maps become more visible. The filtering-based methods show larger deviations from the GT image, whereas the learning-based methods reduce these deviations more effectively. NBGL achieves the highest slice-level PSNR/SSIM of 27.27/0.985. At $\sigma^2$=$0.2$, all methods show increased errors, but NBGL still obtains the best slice-level PSNR/SSIM of 24.87/0.974 among the compared methods. 
The visual error maps and slice-level quantitative results consistently show that NBGL produces outputs closer to the GT image across the three representative noise levels, crucial for preserving the sharpness of the uterine cavity boundary and avoiding downstream clinical misinterpretation caused by tissue over-smoothing.

\begin{table}[!t]
\centering
\footnotesize
    \renewcommand{\arraystretch}{0.7}
\caption{Representative-slice boundary-neighborhood (BN-RMSE) for the qualitative example shown in Figure~\ref{fig:qualitative_comparison_error_maps}. The speckled input is included as a reference, and the best result at each noise level is highlighted in bold.}
\label{tab:target_boundary_neighborhood_rmse}
\begin{tabular*}{0.9\columnwidth}{@{\extracolsep{\fill}}lccc}
\toprule
Method 
& $\sigma^2=0.005$ 
& $\sigma^2=0.05$ 
& $\sigma^2=0.2$ \\
\midrule
Speckled Input
& 0.0299
& 0.0963
& 0.1673 \\
Gaussian Filter
& 0.0426
& 0.0548
& 0.0757 \\
Bilateral Filter
& 0.0365
& 0.0773
& 0.1499 \\
DESD-GAN
& 0.0290
& 0.0542
& 0.0646 \\
DU-GAN
& 0.0434
& 0.0592
& 0.0723 \\
\textbf{NBGL (Ours)}
& \textbf{0.0257}
& \textbf{0.0541}
& \textbf{0.0640} \\
\bottomrule
\end{tabular*}
\end{table}

To further assess reconstruction accuracy near the annotated anatomical boundary in the representative sagittal slice, RMSE was computed within a slice-level boundary-neighborhood region. Specifically, the 2D GT boundary on the selected slice was extracted from the GT boundary map $\mathbf{B}$, dilated using a $3 \times 3$ structuring element for three iterations, and restricted to the annotated region on the same slice. This slice-level measurement was used only to support the qualitative comparison around the annotated anatomical region and is distinct from the full-volume 3D BN-RMSE analysis in Section~\ref{sec:bn_rmse}. 
As shown in Table~\ref{tab:target_boundary_neighborhood_rmse}, NBGL achieves the lowest boundary-neighborhood RMSE across all three representative noise levels. Compared with the speckled input, NBGL reduces this RMSE from 0.0299 to 0.0257 at $\sigma^2$=$0.005$, from 0.0963 to 0.0541 at $\sigma^2$=$0.05$, and from 0.1673 to 0.0640 at $\sigma^2$=$0.2$. 
In addition, under light noise ($\sigma^2$=$0.005$), the Gaussian filter, Bilateral filter, and DU-GAN yield higher boundary-neighborhood RMSE than the speckled input, suggesting that speckle reduction methods without explicit anatomical boundary guidance may introduce additional errors near the boundary, 
primarily due to boundary blurring or over-smoothing of fine structures. 
Together with the slice-level PSNR/SSIM and the visual error maps in Fig.~\ref{fig:qualitative_comparison_error_maps}, these results indicate that NBGL produces despeckled images that better match the GT image while reducing reconstruction errors around the annotated boundary in the representative slice.

\begin{figure}[t]
\setlength{\abovecaptionskip}{0pt}
\setlength{\belowcaptionskip}{0pt}
\setlength{\abovedisplayskip}{0pt}
\setlength{\belowdisplayskip}{0pt}
    \centering    
    \includegraphics[width=\columnwidth]{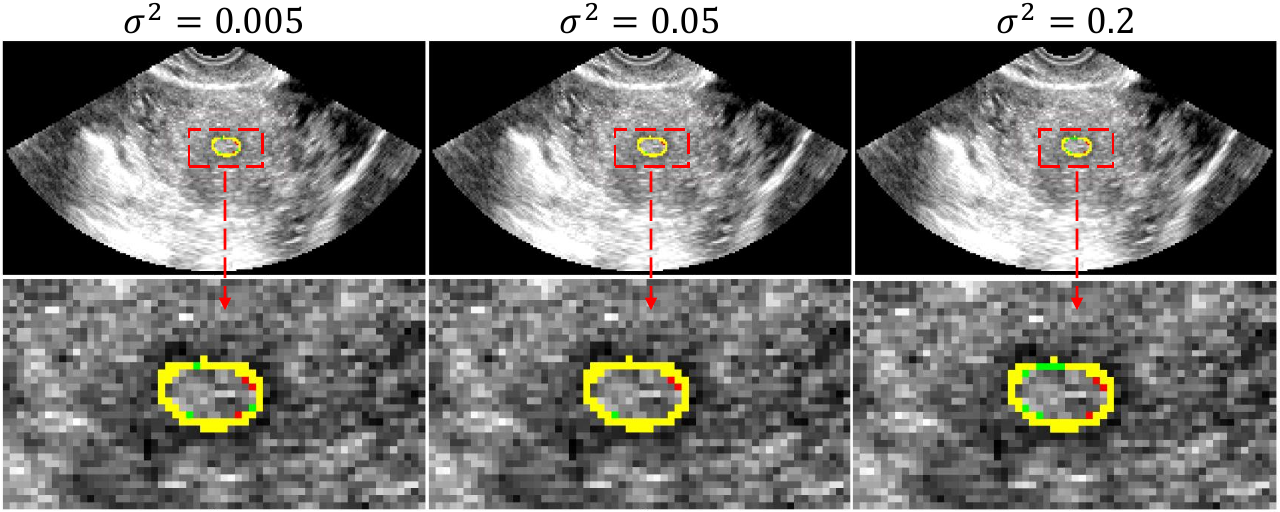}
    \caption{Boundary preservation analysis on a representative sagittal slice under light, moderate, and severe noise levels. Yellow indicates overlapping pixels between the predicted and GT boundaries, green denotes GT-only boundary pixels, and red denotes prediction-only boundary pixels.}    \label{fig:boundary_preservation_analysis}
\end{figure}

\begin{figure*}[!t]
\setlength{\abovecaptionskip}{0pt}
\setlength{\belowcaptionskip}{0pt}
\setlength{\abovedisplayskip}{0pt}
\setlength{\belowdisplayskip}{0pt}
    \centering    
    \includegraphics[width=0.99\textwidth]{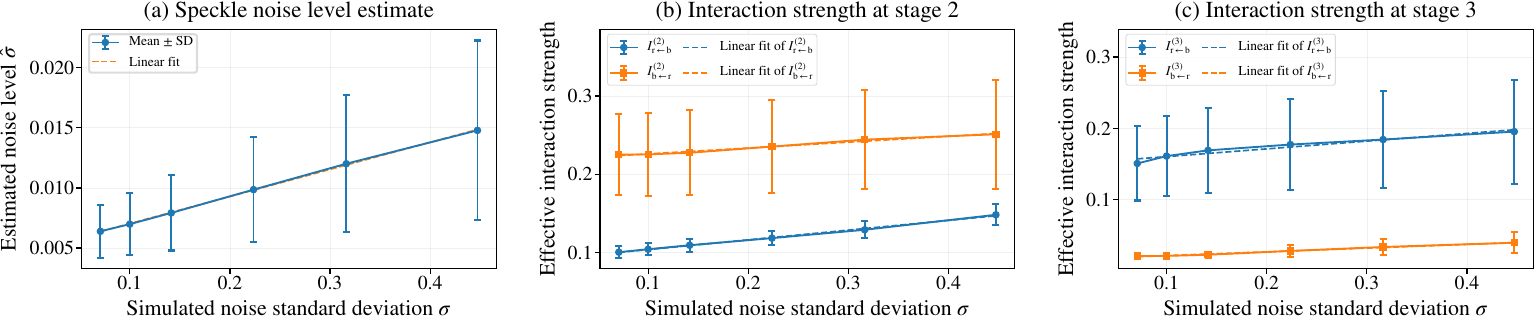}
    \caption{Mechanism analysis across six simulated noise levels on the UterUS test set. (a) MAD-based noise estimate $\hat{\sigma}$ computed during inference. (b,c) Effective cross-branch interaction strengths $I_{\mathrm{r}\leftarrow\mathrm{b}}^{(l)}$ and $I_{\mathrm{b}\leftarrow\mathrm{r}}^{(l)}$ at interaction stages $l=2$ and $l=3$, respectively. Error bars indicate standard deviations across test cases, and dashed lines denote linear fits.}
    \label{fig:mechanism_analysis}
\end{figure*}

\subsection{Boundary Preservation Analysis}
Figure~\ref{fig:boundary_preservation_analysis} compares the predicted boundary with the annotation-derived GT boundary on a representative slice within the annotated region under light ($\sigma^2$=$0.005$), moderate ($\sigma^2$=$0.05$), and severe ($\sigma^2$=$0.2$) noise levels. Yellow indicates overlapping pixels between the predicted and GT boundaries, green denotes GT-only boundary pixels, and red denotes prediction-only boundary pixels. Under light noise, the predicted boundary aligns closely with the GT boundary. Under moderate noise, the boundary remains well aligned with the annotated 
uterine cavity structure, with only minor local deviations.   
Even under severe noise ($\sigma^2$=$0.2$), where local mismatches become more frequent, the overall anatomical geometry and topology are largely preserved without severe distortions. 
This qualitative trend is consistent with both the structural similarity results in Tables~\ref{tab:comparison_low_noise}--\ref{tab:comparison_high_noise} and the slice-level boundary-neighborhood RMSE analysis in Table~\ref{tab:target_boundary_neighborhood_rmse}. 
These observations demonstrate that the proposed boundary enhancement branch successfully guides the network to preserve anatomically meaningful boundaries during speckle reduction.

\subsection{Analysis of Noise-Aware Cross-Branch Feature Coupling}
\label{sec:noise_interaction_mechanism}
To study how NBGL adapts cross-branch feature coupling to varying speckle levels, we analyze the MAD-based speckle estimate $\hat{\sigma}$ used in NIWG and the resulting direction-specific coupling strengths after wFiLM modulation. Ideally, the network should dynamically modulate the cross-branch coupling strength, strengthening guidance from the boundary enhancement branch under severe speckle noise while reducing over-smoothing in low-noise conditions. 
To examine this adaptive behavior in NBGL, 
the MAD-based speckle noise estimates $\hat{\sigma}$ in Eq.~\eqref{eq:mad} and the resulting interaction weights are analyzed across the six simulated log-speckle variance levels, $\sigma^2$$\in$$\{0.005,0.01,0.02,0.05,0.1,0.2\}$. Fig.~\ref{fig:mechanism_analysis} presents the noise estimates obtained during inference and the corresponding direction-specific cross-branch coupling strengths at the two wFiLM interaction stages.

Fig.~\ref{fig:mechanism_analysis}(a) shows the MAD-based noise estimates $\hat{\sigma}$ obtained from the input volume during inference. The mean estimates increase with the simulated noise level, rising from $0.0064\pm0.0022$ at $\sigma^{2}$=$0.005$ to $0.0148\pm0.0075$ at $\sigma^{2}$=$0.2$. 
Although $\hat{\sigma}$ is not intended to provide an exact physical estimate of the simulated speckle variance, it serves as a highly robust, ordered indicator of the input speckle severity. This reliable estimation successfully drives the nonlinear interaction-weight mapping defined in Eq.~\eqref{eq:interaction_weight_mapping}.

Fig.~\ref{fig:mechanism_analysis}(b) and (c) quantify the effective cross-branch interaction strengths at the two interaction stages, $l\in\{2,3\}$. For each direction, the effective interaction strength is measured as the RMS magnitude of the weighted cross-branch modulation term in Eq.~\eqref{eq:film_feature_update}, normalized by the RMS magnitude of the base feature. Specifically, this ratio is defined as 
$I_{t\leftarrow s}^{(l)}=\operatorname{RMS}\left(\omega_{t\leftarrow s}^{(l)}\Delta\mathbf{F}_{t\leftarrow s}^{(l)}\right)/\operatorname{RMS}\left(\mathbf{F}_{t}^{(l)}\right)$, where $s,t\in\{\mathrm{r},\mathrm{b}\}$ and $s\neq t$. Here, $I_{\mathrm{r}\leftarrow\mathrm{b}}^{(l)}$ denotes the interaction from the boundary enhancement branch to the speckle reduction branch at stage $l$, whereas $I_{\mathrm{b}\leftarrow\mathrm{r}}^{(l)}$ denotes the reverse direction.

At Stage~2 (Fig.~\ref{fig:mechanism_analysis}(b)), where the architectural weights are symmetric ($\alpha^{(2)}$=$1$), both $I_{\mathrm{r}\leftarrow\mathrm{b}}^{(2)}$ and $I_{\mathrm{b}\leftarrow\mathrm{r}}^{(2)}$ increase monotonically with the noise level $\sigma$. As $\sigma$ increases from $0.0707$ ($\sigma^{2}$=$0.005$) to $0.4472$ ($\sigma^{2}$=$0.2$), $I_{\mathrm{r}\leftarrow\mathrm{b}}^{(2)}$ increases from $0.1004$$\pm$$0.0076$ to $0.1481$$\pm$$0.0133$, and $I_{\mathrm{b}\leftarrow\mathrm{r}}^{(2)}$ increases from $0.2249$$\pm$$0.0516$ to $0.2510$$\pm$$0.0698$. This monotonic increase confirms that wFiLM amplifies cross-branch coupling under higher estimated noise levels, consistent with Eq.~\eqref{eq:interaction_weight_mapping}.

At Stage~3 (Fig.~\ref{fig:mechanism_analysis}(c)), the asymmetric attenuation factor $\alpha^{(3)}=0.2$ is applied to $\omega_{\mathrm{b}\leftarrow\mathrm{r}}^{(3)}$, resulting in $I_{\mathrm{b}\leftarrow\mathrm{r}}^{(3)}$ being consistently lower than $I_{\mathrm{r}\leftarrow\mathrm{b}}^{(3)}$ across all noise levels. 
Specifically, $I_{\mathrm{r}\leftarrow\mathrm{b}}^{(3)}$ increases from $0.1507\pm0.0519$ to $0.1953\pm0.0728$ as $\sigma^{2}$ increases from $0.005$ to $0.2$, while $I_{\mathrm{b}\leftarrow\mathrm{r}}^{(3)}$ increases from $0.0210\pm0.0029$ to $0.0396\pm0.0146$ over the same range, yielding ratios $I_{\mathrm{b}\leftarrow\mathrm{r}}^{(3)}/I_{\mathrm{r}\leftarrow\mathrm{b}}^{(3)}$ between $0.14$ and $0.20$. 
This reduction reflects the effect of $\alpha^{(3)}$=$0.2$, which attenuates the influence of speckle-reduction features on the boundary representation at the deeper stage while preserving the unattenuated $\omega_{\mathrm{r}\leftarrow\mathrm{b}}^{(3)}$ direction.

The Friedman test confirmed a significant main effect of noise level for all four quantities (all $p$$<$$10^{-12}$, Kendall's $W$ ranging from $0.474$ to $1.000$). 
Based on this overall effect, post hoc pairwise comparisons were performed and are reported in Supplementary Table~SII. 
The two quantities $I_{\mathrm{r}\leftarrow\mathrm{b}}^{(2)}$ and $I_{\mathrm{r}\leftarrow\mathrm{b}}^{(3)}$ showed perfect rank concordance ($\chi^{2}(5)=140.00$, $W=1.000$), whereas $I_{\mathrm{b}\leftarrow\mathrm{r}}^{(2)}$ and $I_{\mathrm{b}\leftarrow\mathrm{r}}^{(3)}$ showed moderate concordance ($I_{\mathrm{b}\leftarrow\mathrm{r}}^{(2)}$: $W$=$0.474$; $I_{\mathrm{b}\leftarrow\mathrm{r}}^{(3)}$: $W$=$0.603$). Post hoc Holm-corrected Wilcoxon signed-rank tests showed that all $15$ pairwise differences were significant for $I_{\mathrm{r}\leftarrow\mathrm{b}}^{(2)}$ and $I_{\mathrm{r}\leftarrow\mathrm{b}}^{(3)}$, with all corrected $p$-values below $0.001$, indicating that the interaction from the boundary enhancement branch to the speckle reduction branch changed consistently with the noise level. In contrast, the reverse interaction showed weaker sensitivity at lower noise levels. For $I_{\mathrm{b}\leftarrow\mathrm{r}}^{(2)}$, only the comparison between the two lowest noise levels ($\sigma^{2}$=$0.005$ vs. $0.010$) was not significant, whereas all other pairwise differences were significant after Holm correction. For $I_{\mathrm{b}\leftarrow\mathrm{r}}^{(3)}$, the comparisons of $\sigma^{2}$=$0.005$ with $0.010$ and $0.020$ were not significant, whereas all remaining pairwise differences were significant after Holm correction. These results indicate that the interaction from the boundary enhancement branch to the speckle reduction branch is more sensitive to increasing noise levels, whereas the reverse interaction changes more gradually at lower noise levels.

These results are consistent with the design of the noise-aware interaction mechanism. The wFiLM module increases cross-branch coupling strength in response to higher estimated noise levels at both interaction stages. 
Meanwhile, the asymmetric attenuation $\alpha^{(3)}$=$0.2$ at stage~3 limits the influence of speckle-reduction features on the boundary representation while preserving the unattenuated $\omega_{\mathrm{r}\leftarrow\mathrm{b}}^{(3)}$ direction.

\begin{table}[!t]
    \centering
    \renewcommand{\arraystretch}{0.7}  
    \caption{Ablation study of the major loss terms and architectural components in NBGL on the mixed-noise UterUS test set. The best results are highlighted in bold.}
    \label{tab:ablation}
    \resizebox{\linewidth}{!}{%
    \begin{tabular}{l|ccc}
        \toprule
        Method 
        & PSNR$\uparrow$ 
        & SSIM$\uparrow$ 
        & RMSE$\downarrow$  \\
        \midrule
        {w/o $\mathcal{L}_{B}$}
        & {26.6300}{$\pm$1.8298}
        & {0.9442}{$\pm$0.0174}
        & {0.0501}{$\pm$0.0099}  \\
        {w/o $\mathcal{L}_{S}$}
        & {26.5971}{$\pm$1.7976}
        & {0.9393}{$\pm$0.0203}
        & {0.0502}{$\pm$0.0099}  \\      
        {w/o $\mathcal{L}_{S}$$\mathcal{L}_{B}$}
        & {26.3435}{$\pm$1.8589}
        & {0.9377}{$\pm$0.0195}
        & {0.0518}{$\pm$0.0106} \\
        \midrule
        {w/o Boundary Branch}
        & {27.0098}{$\pm$1.9013}
        & {0.9495}{$\pm$0.0168}
        & {0.0481}{$\pm$0.0101} \\
        {w/o wFiLM}
        & {26.9689}{$\pm$1.8889}
        & {0.9493}{$\pm$0.0167}
        & {0.0484}{$\pm$0.0101} \\
        {Additive Interaction}
        & {24.7771}{$\pm$1.9999}
        & {0.9266}{$\pm$0.0244}
        & {0.0648}{$\pm$0.0154} \\
        \midrule
        \textbf{{NBGL (Ours)}}
        & \textbf{{27.0587}{$\pm$1.8528}}
        & \textbf{{0.9498}{$\pm$0.0165}}
        & \textbf{{0.0479}{$\pm$0.0098}} \\
        \bottomrule
    \end{tabular}
    } 
\end{table}

\section{Discussion}
\label{sec:discussion}

\subsection{Ablation Study}
To examine the contribution of each major component in NBGL, ablation experiments are conducted on the mixed-noise UterUS test set. The evaluated variants cover two aspects: loss-term ablations based on the overall objective in Eq.~\eqref{eq:overall_loss}, and architectural ablations involving the boundary enhancement branch and wFiLM-based cross-branch feature coupling. The quantitative results are summarized in Table~\ref{tab:ablation}.

The first group of ablations evaluates the contribution of individual loss terms in the overall objective in Eq.~\eqref{eq:overall_loss}. To assess the effect of boundary-specific supervision, the variant ``\textbf{w/o $\mathcal{L}_{B}$}''  removes the boundary preservation loss during training while retaining the remaining terms. This variant achieves lower performance than NBGL across all three metrics, with a PSNR of 26.6300, an SSIM of 0.9442, and an RMSE of 0.0501, corresponding to decreases of 0.4287 in PSNR and 0.0056 in SSIM, and an increase of 0.0022 in RMSE compared with NBGL. 
The variant ``\textbf{w/o $\mathcal{L}_{S}$}'' excludes the 3D SSIM loss while keeping the remaining loss terms unchanged. This variant causes a larger SSIM drop than removing $\mathcal{L}_{B}$ alone, with a decrease of 0.0105 compared with NBGL, indicating that the structural similarity loss term is important for maintaining global structural consistency during speckle reduction.   
Similarly, the variant ``\textbf{w/o $\mathcal{L}_{S}\mathcal{L}_{B}$}''  
excludes the 3D SSIM loss and the boundary preservation loss. 
The resulting performance degradation validates that structural similarity supervision and boundary-specific supervision jointly benefit the preservation of fine anatomical details and organic textures. 

The second group evaluates the efficacy of our core architectural designs.  
The variant ``\textbf{w/o Boundary Branch}'' completely detaches the auxiliary boundary stream, reducing the network to a single-stream denoising model. 
As shown in Table~\ref{tab:ablation}, this configuration leads to a slight performance decline, demonstrating that the boundary enhancement branch provides useful  structural priors. 
Furthermore, the variant ``\textbf{w/o wFiLM}'' keeps the dual-branch architecture but removes explicit wFiLM-based cross-branch feature modulation. 
The performance drop of this variant proves that explicit cross-branch feature coupling is beneficial for handling complex noise variations. 
Lastly, the variant ``\textbf{Additive Interaction}'' substitutes the modulated features with a basic element-wise addition. The superior performance of full NBGL over this variant shows that wFiLM enables more effective adaptive feature regulation than simple additive fusion.

\subsection{Parameter Sensitivity Analysis}
Hyperparameter sensitivity was evaluated on the mixed-noise UterUS test set. The analysis includes the power-law exponent $\rho$ in the interaction-weight mapping (Eq.~\eqref{eq:interaction_weight_mapping}), the direction-specific attenuation factor $\alpha^{(3)}$ at the deepest interaction stage (Eq.~\eqref{eq:dynamic_interaction_weights}), and the loss weights $\lambda_{S}$ and $\lambda_{B}$ in Eq.~\eqref{eq:overall_loss}. Results are summarized in Tables~\ref{tab:rho_analysis}--\ref{tab:sensitivity_boundary}.

\begin{table}[!t]
    \renewcommand{\arraystretch}{0.7}
    \centering
    \scriptsize
    \caption{Effect of the power-law exponent $\rho$ on the mixed-noise UterUS test set. The best results are highlighted in bold.}
    \label{tab:rho_analysis}
    \resizebox{\linewidth}{!}{%
    \begin{tabular}{l|cccc}
        \toprule
        $\rho$ 
        & PSNR$\uparrow$ 
        & SSIM$\uparrow$ 
        & RMSE$\downarrow$ \\
        \midrule
        $\rho=1.0$
        & {26.8352}{$\pm$1.7909}
        & {0.9491}{$\pm$0.0164}
        & {0.0490}{$\pm$0.0096} \\
        $\rho=1.5$
        & {26.9926}{$\pm$1.8408}
        & \textbf{{0.9501}{$\pm$0.0163}}
        & {0.0483}{$\pm$0.0099} \\
        $\bm{\rho=2.0}$
        & \textbf{{27.0587}{$\pm$}1.8528}
        & {0.9498}{$\pm$0.0165}
        & \textbf{{0.0479}{$\pm$0.0098}} \\
        $\rho=2.5$
        & {27.0119}{$\pm$1.9061}
        & {0.9493}{$\pm$0.0165}
        & {0.0481}{$\pm$0.0101} \\
        $\rho=3.0$
        & {27.0177}{$\pm$1.8310}
        & {0.9499}{$\pm$0.0164}
        & {0.0481}{$\pm$0.0097} \\
        \bottomrule
    \end{tabular}
    } 
\end{table}

\subsubsection{Effect of Nonlinear Exponent}
The sensitivity analysis results regarding the power-law exponent $\rho$ in Eq.~\eqref{eq:interaction_weight_mapping} are summarized in Table~\ref{tab:rho_analysis}. Compared with the baseline linear mapping ($\rho = 1.0$), all nonlinear configurations ($\rho > 1.0$) consistently yield superior quantitative metrics. Specifically, incorporating a moderate nonlinearity with $\rho = 1.5$ increases PSNR from 26.8352 to 26.9926 and SSIM from 0.9491 to 0.9501. 
Our default setting ($\rho = 2.0$) achieves the optimal overall performance, reaching the highest PSNR of 27.0587 and the lowest RMSE of 0.0479, which represent an absolute improvement of 0.2235 in PSNR and a reduction of 0.0011 in RMSE over the linear baseline. When the exponent is further increased to $\rho = 2.5$ and $\rho = 3.0$, the performance gains slightly plateau or degrade marginally in terms of PSNR and RMSE, though they still outperform the linear case. Compared to the optimal $\rho = 2.0$, $\rho = 1.5$ and $\rho = 3.0$ exhibit a slight decrease in PSNR (by 0.0661 and 0.0410, respectively), despite capturing marginally higher SSIM values. 
These comprehensive comparisons demonstrate that a nonlinear weight mapping is better suited for adaptively regulating cross-branch feature coupling across fluctuating noise levels. By keeping the interaction weights relatively suppressed under low-noise conditions and amplifying them nonlinearly under severe noise, the network effectively leverages boundary-aware guidance without suffering from tissue over-smoothing. Overall, $\rho = 2.0$ provides the best performance balance while ensuring robust stability under reasonable variations of the exponent.

\subsubsection{Effect of Asymmetric Interaction Factor}
The sensitivity to the direction-specific attenuation factor $\alpha^{(3)}$ at the deepest interaction stage in Eq.~\eqref{eq:dynamic_interaction_weights} is summarized in Table~\ref{tab:alpha3_analysis}. 
This factor controls the feedback intensity from the speckle reduction branch back to the boundary enhancement branch within the deep latent space. 
When this feedback path is completely deactivated ($\alpha^{(3)}$=$0$), the model experiences a prominent performance decline compared with the optimal setting, with the PSNR dropping by 0.3606 (from 27.0587 to 26.6981) and the SSIM decreasing by 0.0025, while the RMSE increases by 0.0018. This severe degradation implies that the deep semantic features extracted by the speckle reduction branch contain rich contextual information that remains highly beneficial for guiding robust boundary feature extraction. 
Our default non-zero attenuated setting ($\alpha^{(3)}$=$0.2$) achieves the best overall performance across all three quantitative metrics. When the factor is increased to a moderate level ($\alpha^{(3)}$=$0.5$), the metrics degrade slightly, with a decrease of 0.0365 in PSNR and an increase of 0.0002 in RMSE. Crucially, when transitioning to the fully symmetric configuration ($\alpha^{(3)}$=$1.0$), we observe a much more pronounced and consistent performance penalty, where the PSNR drops by 0.0803, and the SSIM decreases by 0.0005. 
These comprehensive comparisons validate our core design intuition: while deep cross-branch feature sharing is beneficial, a fully symmetric interaction allows unfiltered noise residuals from the speckle reduction branch to back-propagate into and contaminate the boundary enhancement stream. Consequently, implementing an asymmetric attenuation via $\alpha^{(3)}$=$0.2$ provides the optimal trade-off, effectively borrowing advanced contextual priors from the denoising branch while insulating the boundary representations from noise contamination. 

\begin{table}[!t]
    \renewcommand{\arraystretch}{0.7}
    \scriptsize
    \centering
    \caption{Effect of the asymmetric interaction factor $\alpha^{(3)}$ on the mixed-noise UterUS test set. The best results are highlighted in bold.}
    \label{tab:alpha3_analysis}
    \resizebox{\linewidth}{!}{%
    \begin{tabular}{l|cccc}
        \toprule
        $\alpha^{(3)}$ 
        & PSNR$\uparrow$ 
        & SSIM$\uparrow$ 
        & RMSE$\downarrow$ \\
        \midrule
        $\alpha^{(3)}=0.0$
        & {26.6981}{$\pm$1.8511}
        & {0.9473}{$\pm$0.0170}
        & {0.0497}{$\pm$0.0100} \\
        $\bm{\alpha^{(3)}=0.2}$
        & \textbf{{27.0587}{$\pm$1.8528}}
        & \textbf{{0.9498}{$\pm$0.0165}}
        & \textbf{{0.0479}{$\pm$0.0098}} \\
        $\alpha^{(3)}=0.5$
        & {27.0222}{$\pm$1.8915}
        & {0.9497}{$\pm$0.0167}
        & {0.0481}{$\pm$0.0101} \\
        $\alpha^{(3)}=1.0$
        & {26.9784}{$\pm$1.8694}
        & {0.9493}{$\pm$0.0165}
        & {0.0484}{$\pm$0.0099} \\
        \bottomrule
    \end{tabular}
    } 
\end{table}

\subsubsection{Effect of Key Parameters}
\label{sec:loss_weight}
To evaluate the impact of joint loss regularization, we analyze the sensitivity of the network to two critical hyperparameters: the structural similarity loss weight $\lambda_S$ (in the speckle reduction objective) and the boundary preservation loss weight $\lambda_B$ (in the overall objective). The quantitative evaluations are summarized in Table~\ref{tab:sensitivity_structural} and Table~\ref{tab:sensitivity_boundary}, respectively. 

Table~\ref{tab:sensitivity_structural}  outlines the performance fluctuations across different values of $\lambda_S$. 
Our default configuration ($\lambda_S $=$ 4.0$) achieves the optimal overall balance, yielding the highest PSNR of 27.0587 and the lowest RMSE of 0.0479. 
While a slightly lower weight ($\lambda_S $=$3.0$) captures a marginally higher SSIM (0.9499 vs. 0.9498), it compromises the intensity reconstruction accuracy, decreasing PSNR by 0.1258 and increasing RMSE by 0.0008. Reducing $\lambda_S$ further (to 0.5, 1.0, and 2.0) leads to consistent performance degradation across all metrics. 
Conversely, an excessively large weight ($\lambda_S$=$5.0$) also reduces PSNR by 0.0896. 
This behavior indicates that while strong structural constraints are essential to suppress severe speckle artifacts, overly aggressive optimization on SSIM may over-smooth fine-grained intensity variations, thereby plateauing the reconstruction quality.

The variations across different boundary preservation loss weights $\lambda_B$ are reported in Table~\ref{tab:sensitivity_boundary}. The default setting ($\lambda_B$=$0.5$) consistently outperforms all alternative configurations. 
Reducing the boundary supervision ($\lambda_B$=$0.1$) results in a drop of 0.1665 in PSNR and 0.0011 in SSIM, confirming that insufficient boundary guidance weakens the framework's ability to capitalize on anatomical priors. 
On the other hand, increasing $\lambda_B$ beyond 0.5 introduces a sharp performance penalty; for instance, $\lambda_B$=$3.0$ and $4.0$ decrease PSNR by 0.2838 and 0.2988, respectively, while increasing RMSE. 
This trend validates our design intuition: moderate boundary supervision successfully guides feature representation during speckle reduction, whereas an excessively large $\lambda_B$ biases the joint optimization heavily toward edge detection, distracting the network from its primary task of intensity denoising and tissue reconstruction.

\begin{table}[!t]
    \renewcommand{\arraystretch}{0.7}
    \scriptsize    
    \centering
    \caption{Effect of the structural similarity loss weight $\lambda_{S}$ on the mixed-noise UterUS test set. The best results are highlighted in bold.}
    \label{tab:sensitivity_structural}
    \resizebox{\linewidth}{!}{%
    \begin{tabular}{l|cccc}
        \toprule
        $\lambda_{S}$  
        & PSNR$\uparrow$ 
        & SSIM$\uparrow$ 
        & RMSE$\downarrow$  \\
        \midrule
        $\lambda_{S}=0.5$
        & {26.9432}{$\pm$1.9201}
        & {0.9466}{$\pm$0.0180}
        & {0.0483}{$\pm$0.0101} \\
        $\lambda_{S}=1.0$
        & {26.8710}{$\pm$1.8184}
        & {0.9488}{$\pm$0.0165}
        & {0.0488}{$\pm$0.0098} \\
        $\lambda_{S}=2.0$
        & {26.9646}{$\pm$1.8889}
        & {0.9488}{$\pm$0.0166}
        & {0.0485}{$\pm$0.0101} \\
        $\lambda_{S}=3.0$
        & {26.9329}{$\pm$1.8934}
        &  \textbf{{0.9499}{$\pm$0.0165}}
        & {0.0487}{$\pm$0.0102} \\
        $\bm{\lambda_{S}=4.0}$
        & \textbf{{27.0587}{$\pm$1.8528}}
        &{0.9498}{$\pm$0.0165}
        & \textbf{{0.0479}{$\pm$0.0098}} \\
        $\lambda_{S}=5.0$
        & {26.9691}{$\pm$1.8311}
        & {0.9498}{$\pm$0.0162}
        & {0.0484}{$\pm$0.0097} \\
        \bottomrule
    \end{tabular}
    } 
\end{table}

\begin{table}[!t]
    \renewcommand{\arraystretch}{0.7}
    \scriptsize
    \centering
    \caption{Effect of the boundary preservation loss weight $\lambda_{B}$ on the mixed-noise UterUS test set. The best results are highlighted in bold.}
    \label{tab:sensitivity_boundary}
    \resizebox{\linewidth}{!}{%
    \begin{tabular}{l|cccc}
        \toprule
        $\lambda_{B}$ 
        & PSNR$\uparrow$ 
        & SSIM$\uparrow$ 
        & RMSE$\downarrow$ \\
        \midrule
        $\lambda_{B}=0.1$
        &{26.8922}{$\pm$1.8597}
        & {0.9487}{$\pm$0.0165}
        & {0.0487}{$\pm$0.0099} \\
        $\bm{\lambda_{B}=0.5}$
        & \textbf{{27.0587}{$\pm$1.8528}}
        & \textbf{{0.9498}{$\pm$0.0165}}
        & \textbf{{0.0479}{$\pm$0.0098}} \\
        $\lambda_{B}=1.0$
        & {26.8982}{$\pm$1.8692}
        & {0.9491}{$\pm$0.0168}
        & {0.0488}{$\pm$0.0100} \\
        $\lambda_{B}=2.0$
        & {26.8802}{$\pm$1.8569}
        & {0.9494}{$\pm$0.0166}
        & {0.0488}{$\pm$0.0099} \\
         $\lambda_{B}=3.0$
        & {26.7749}{$\pm$1.8649}
        & {0.9487}{$\pm$0.0170}
        & {0.0495}{$\pm$0.0100} \\
        $\lambda_{B}=4.0$
        & {26.7599}{$\pm$1.8692}
        & {0.9482}{$\pm$0.0169}
        & {0.0494}{$\pm$0.0100} \\
        $\lambda_{B}=5.0$
        & {26.8145}{$\pm$1.7375}
        & {0.9493}{$\pm$0.0169}
        & {0.0491}{$\pm$0.0095} \\
        \bottomrule
    \end{tabular}
    } 
\end{table}

\subsection{Effect of Boundary Guidance on Speckle Reduction}
A central component of the NBGL framework is the explicit integration of annotation-derived anatomical boundary supervision through an auxiliary boundary enhancement branch. 
This design is motivated by the observation that optimizing ultrasound despeckling models solely for intensity reconstruction can degrade subtle yet clinically important anatomical boundaries, particularly when speckle noise is severe.
Although previous gradient- or edge-guided methods have been explored for structural preservation~\citep{zhou2024gradient, huang2021gan, han2022dual}, relying only on local gradient or edge responses may be insufficient when severe speckle noise obscures tissue boundaries. 
In contrast, in NBGL, the speckle reduction branch focuses on reconstructing speckle-reduced intensities, while the boundary enhancement branch learns boundary-sensitive representations under supervision from boundaries derived from annotation masks.
Through noise-aware cross-branch interaction, these boundary-sensitive representations are incorporated into the speckle reduction branch, helping preserve clinically relevant anatomical boundaries while reducing speckle.

This interpretation is supported by  quantitative and qualitative results. 
In the mixed-noise evaluation (Table~\ref{tab:comparison_mixed}), NBGL achieves the highest mean SSIM (0.9498) with a relatively small standard deviation of 0.0165, indicating stable structural similarity under mixed-noise conditions. The per-noise-level results in Tables~\ref{tab:comparison_low_noise}--\ref{tab:comparison_high_noise} also show that NBGL maintains high SSIM across all evaluated noise levels. 
In addition, the ablation results in Table~\ref{tab:ablation} show that removing $\mathcal{L}_{B}$, jointly removing $\mathcal{L}_{S}$ and $\mathcal{L}_{B}$, or removing the boundary enhancement branch leads to lower SSIM, indicating that both boundary-related supervision and the auxiliary boundary branch contribute to structural preservation. As shown in Fig.~\ref{fig:qualitative_comparison_error_maps}, NBGL produces despeckled images that better match the GT image across light, moderate, and severe noise levels. The slice-level boundary-neighborhood RMSE in 
   Table~\ref{tab:target_boundary_neighborhood_rmse} shows that NBGL achieves the lowest boundary-neighborhood error across all three representative noise levels, including $\sigma^2$=$0.005$, $\sigma^2$=$0.05$, and $\sigma^2$=$0.2$. 
The boundary preservation analysis in Fig.~\ref{fig:boundary_preservation_analysis} further shows that the predicted boundaries remain close to the GT boundaries under light and moderate noise. Under severe noise, local mismatches become more frequent, but the overall anatomical shape remains largely preserved. 
These findings suggest that the boundary enhancement branch, supported by explicit boundary supervision, contributes to preserving annotation-derived anatomical boundaries during speckle reduction.

\subsection{Effect of Noise-Aware Coupling on Speckle Reduction}
Beyond boundary guidance, NBGL adapts cross-branch feature coupling to the estimated input speckle noise level, improving the robustness of speckle reduction across varying noise conditions.
NIWG estimates the speckle noise level from the input volume using 3D Laplacian filtering and MAD estimation, and maps the estimate to an interaction weight. 
The resulting weight is then incorporated into wFiLM to modulate bidirectional feature coupling between the speckle reduction branch and the boundary enhancement branch. Under lower noise levels, the cross-branch coupling remains relatively weak, whereas under higher noise levels, stronger coupling allows boundary-aware features to contribute more to speckle reduction.

This behavior is supported by the internal analysis in Fig.~\ref{fig:mechanism_analysis}. The MAD-based noise estimates obtained during inference increase with the simulated noise level, providing an ordered input for the NIWG mapping. The effective interaction strengths $I_{\mathrm{r}\leftarrow\mathrm{b}}^{(2)}$, $I_{\mathrm{r}\leftarrow\mathrm{b}}^{(3)}$, $I_{\mathrm{b}\leftarrow\mathrm{r}}^{(2)}$, and $I_{\mathrm{b}\leftarrow\mathrm{r}}^{(3)}$ also increase with noise level, but with different sensitivities across directions and stages. 
As analyzed in Section~\ref{sec:noise_interaction_mechanism}, $I_{\mathrm{r}\leftarrow\mathrm{b}}^{(2)}$ and $I_{\mathrm{r}\leftarrow\mathrm{b}}^{(3)}$ show significant pairwise differences across all noise-level pairs, indicating that the interaction from the boundary enhancement branch to the speckle reduction branch changes consistently with increasing noise. In contrast, $I_{\mathrm{b}\leftarrow\mathrm{r}}^{(2)}$ and $I_{\mathrm{b}\leftarrow\mathrm{r}}^{(3)}$ show fewer significant differences at low noise levels, indicating that the reverse interaction changes less in the low-noise range. This direction-specific behavior is consistent with the design in Eq.~\eqref{eq:dynamic_interaction_weights}, where the reverse interaction at the deepest stage is attenuated by $\alpha^{(3)}$=$0.2$.

The contribution of this noise-aware mechanism is further supported by the ablation and sensitivity results. As shown in Table~\ref{tab:ablation}, removing wFiLM lowers performance, and replacing wFiLM with Additive Interaction leads to a larger performance drop. These results indicate that simple feature addition is less effective than adaptive wFiLM-based modulation for cross-branch coupling. The sensitivity analyses in Tables~\ref{tab:rho_analysis} and \ref{tab:alpha3_analysis} further show that NBGL remains numerically stable under moderate variations of $\rho$ and $\alpha^{(3)}$. 
These results suggest that the NIWG-wFiLM mechanism improves robustness to heterogeneous noise levels by adjusting cross-branch feature coupling according to the estimated input speckle level.

\subsection{Generalization to Unseen Noise Levels}
To evaluate the generalization capability of the evaluated models to noise conditions outside the mixed-noise training distribution, all methods are tested across three unseen log-speckle variance levels: $\sigma^2 \in \{0.003, 0.03, 0.3\}$. 
This selection spans a comprehensive test bed including one ultra-light noise level below the training range ($\sigma^2$=$0.003$), one intermediate level within the range but unsampled during training ($\sigma^2$=$0.03$), and one severe noise level shifting beyond the training boundaries ($\sigma^2$=$0.3$).   
Quantitative results are summarized in Table~\ref{tab:comparison_unseen_mixed_noise}. Detailed results for individual unseen noise levels are provided in Supplementary Table~SIII.

At the \emph{ultra-light noise level} ($\sigma^2$=$0.003$), traditional filtering-based methods remain highly competitive. Notably, the patch-based BM4D baseline achieves an impressive PSNR of 32.0092 and an SSIM of 0.9871, outperforming the strongest learning-based baseline (DESD-GAN) by 0.9079 in PSNR. This trend indicates that non-local patch recurring algorithms preserve fine textures exceptionally well when speckle artifacts are minimal. Nevertheless, NBGL consistently outperforms both paradigms, establishing the top performance with a PSNR of 33.0364 and an SSIM of 0.9897. This corresponds to an absolute improvement of 1.0272 in PSNR over BM4D and 1.9351 over DESD-GAN, demonstrating that NBGL effectively scales down to pristine imaging conditions without introducing unwanted network-generated artifacts or blurriness. 

At the \emph{intermediate unseen noise level} ($\sigma^2$=$0.03$), a distinct paradigm shift occurs where deep learning baselines begin to significantly outpace filtering methods. Among the competitors, DESD-GAN yields the strongest baseline metrics (PSNR = 26.2557, SSIM = 0.9456). NBGL further improves upon these results, achieving the highest performance across all three indicators (PSNR = 26.8547, SSIM = 0.9573, RMSE = 0.0464). The steady margin over unsampled interpolation noise levels validates the robustness of the interpolation provided by our noise-aware interaction weight generation module. 

At the \emph{severe unseen noise level} ($\sigma^2$=$0.3$), the heavy speckle corruptions severely degrade the intensity profiles, causing traditional filters to plateau early. Here, DESD-GAN remains our closest competitor (PSNR = 22.2191, RMSE = 0.0787). While NBGL yields a moderate gain in terms of absolute intensity metrics (improving PSNR by 0.2134 and reducing RMSE by 0.0015 over DESD-GAN), it achieves a substantial and statistically significant leap in structural preservation, boosting the SSIM by 0.0204 (reaching 0.8852). This critical observation highlights that under extreme, out-of-distribution noise where intensity profiles are deeply obliterated, the explicit guidance from our expert-knowledge boundary stream serves as a structural anchor, successfully capturing macroscopic anatomy where standard generative methods fail.

Aggregated across all unseen configurations, NBGL achieves optimal performance across the entire test spectrum (Mean PSNR=27.4412, SSIM=0.9441, RMSE=0.0488), yielding substantial margins over the top filtering baseline (Gaussian Filter) and the deep learning baseline (DESD-GAN). 
The Holm-adjusted $p$-values ($p_{\text{Holm}}$=$8.94 \times 10^{-8}$) in Table~\ref{tab:comparison_unseen_mixed_noise} confirm that the structural similarity improvements achieved by NBGL are statistically significant across the board, guaranteeing robust generalization to unpredictable clinical ultrasound variations.

\begin{table}[!t]
    \renewcommand{\arraystretch}{0.7}
    \scriptsize
    \centering
    \caption{Quantitative comparison of the proposed NBGL against 12 baseline models for ultrasound speckle reduction on the UterUS test set across all unseen noise levels. The best results are highlighted in bold; $p_{\mathrm{Holm}}$ denotes Holm-adjusted $p$-values.}    
    \label{tab:comparison_unseen_mixed_noise}
    \resizebox{\linewidth}{!}{%
    \begin{tabular}{l|cccc}
        \toprule  
        \multirow{2}{*}{Method}
        & \multicolumn{4}{c}{All unseen noise levels} \\
        \cmidrule(lr){2-5}
        & PSNR$\uparrow$ 
        & SSIM$\uparrow$ 
        & RMSE$\downarrow$ 
        & $p_{\mathrm{Holm}}$ \\
        \midrule
        Gaussian Filter
        & {23.1644}{$\pm$1.9153}
        & {0.8886}{$\pm$0.0216}
        & {0.0745}{$\pm$0.0164}
        & {8.94$\times 10^{-8}$} \\
        Mean Filter
        & {21.4691}{$\pm$1.9122}
        & {0.8227}{$\pm$0.0312}
        & {0.0874}{$\pm$0.0187}
        & {8.94$\times 10^{-8}$} \\
        Median Filter
        & {21.3528}{$\pm$1.6684}
        & {0.8317}{$\pm$0.0258}
        & {0.0898}{$\pm$0.0166}
        & {8.94$\times 10^{-8}$} \\
        Bilateral Filter
        & {22.3716}{$\pm$1.3555}
        & {0.8764}{$\pm$0.0173}
        & {0.0985}{$\pm$0.0172}
        & {8.94$\times 10^{-8}$} \\
        BM4D
        & {22.7256}{$\pm$1.8403}
        & {0.8844}{$\pm$0.0254}
        & {0.1034}{$\pm$0.0197}
        & {8.94$\times 10^{-8}$} \\
        \midrule
        DESD-GAN
        & {26.5254}{$\pm$2.0060}
        & {0.9308}{$\pm$0.0222}
        & {0.0525}{$\pm$0.0108}
        & {8.94$\times 10^{-8}$} \\
        DU-GAN
        & {25.5223}{$\pm$1.1269}
        & {0.9181}{$\pm$0.0166}
        & {0.0597}{$\pm$0.0089}
        & {8.94$\times 10^{-8}$} \\
        G2CR-FPN
        & {23.0540}{$\pm$1.3407}
        & {0.7196}{$\pm$0.0433}
        & {0.0724}{$\pm$0.0114}
        & {8.94$\times 10^{-8}$} \\
        LIT-Former
        & {10.7119}{$\pm$1.9587}
        & {0.4997}{$\pm$0.0357}
        & {0.2986}{$\pm$0.0677}
        & {8.94$\times 10^{-8}$} \\
        Speckle2Self
        & {17.0373}{$\pm$2.2717}
        & {0.6489}{$\pm$0.0345}
        & {0.1456}{$\pm$0.0403}
        & {8.94$\times 10^{-8}$} \\
        Autoencoder-KL
        & {19.4773}{$\pm$1.0745}
        & {0.5687}{$\pm$0.1612}
        & {0.1071}{$\pm$0.0133}
        & {8.94$\times 10^{-8}$} \\
        DDIM
        & {15.4070}{$\pm$3.2743}
        & {0.5245}{$\pm$0.1278}
        & {0.1832}{$\pm$0.0728}
        & {8.94$\times 10^{-8}$} \\
        \midrule
        \textbf{NBGL (Ours)}
        & \textbf{{27.4412}{$\pm$1.8992}}
        & \textbf{{0.9441}{$\pm$0.0168}}
        & \textbf{{0.0488}{$\pm$0.0102}}
        & \textbf{--} \\
        \bottomrule
    \end{tabular}
    }
\end{table}

\begin{figure*}[!h]
\setlength{\abovecaptionskip}{0pt}
\setlength{\belowcaptionskip}{-4pt}
\setlength{\abovedisplayskip}{0pt}
\setlength{\belowdisplayskip}{0pt}
    \centering
    \includegraphics[width=1\textwidth]{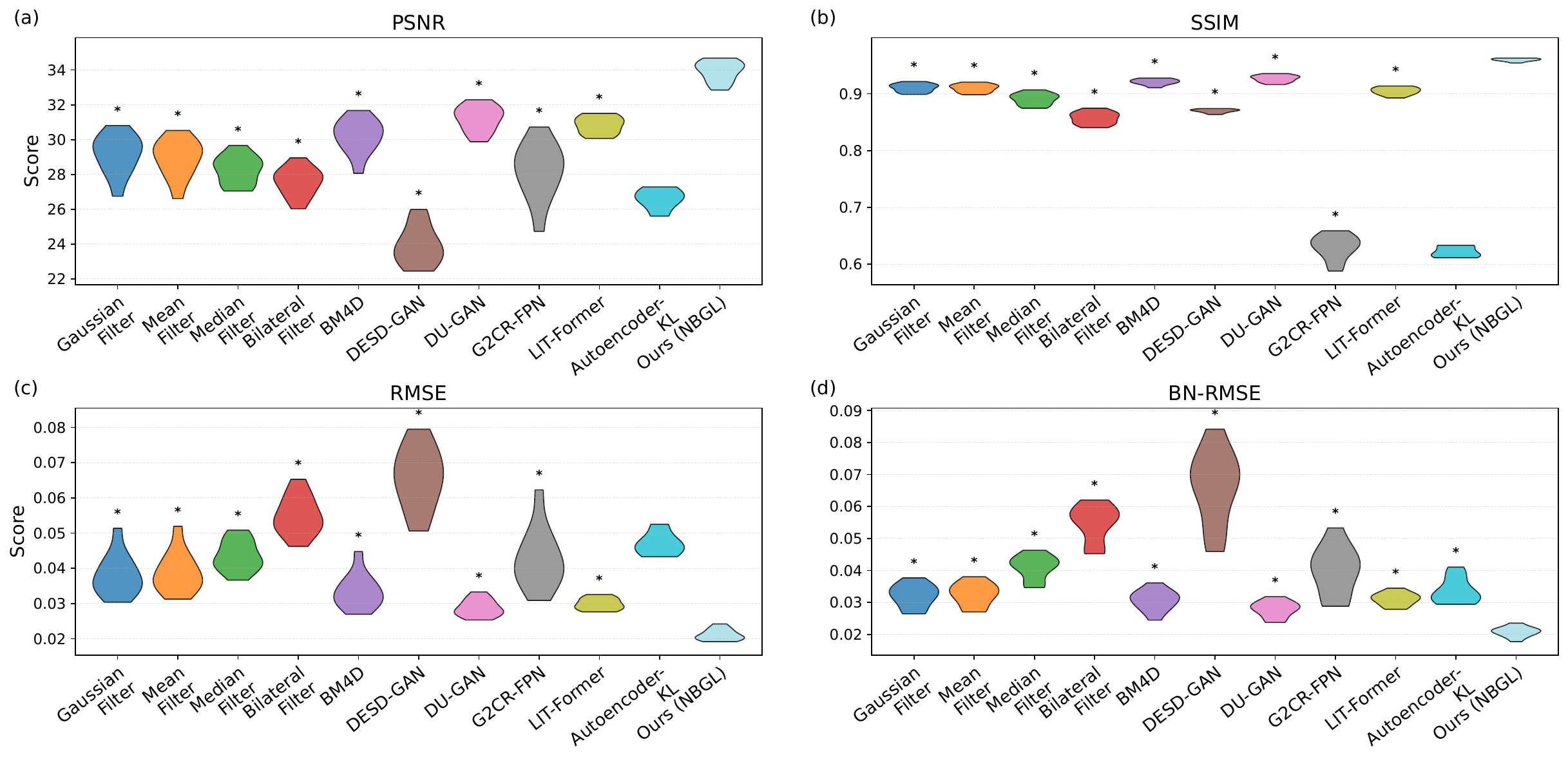}
    \caption{Quantitative evaluation on the TRUSTED test set. The distributions of PSNR, SSIM, RMSE, and BN-RMSE are shown after aggregating the results over the six evaluated speckle noise levels. 
    Statistical significance between NBGL and each competing method was assessed separately for each metric using paired Wilcoxon signed-rank tests with Holm correction over all baseline comparisons, where * indicates $p_{\mathrm{Holm}}<0.05$.}   
    \label{fig:trusted_quantitative}
\end{figure*}

\subsection{Boundary-Neighborhood Reconstruction Accuracy}
\label{sec:bn_rmse}
To further evaluate the reconstruction fidelity specifically near the annotation-derived anatomical landmarks, we compute the boundary-neighborhood RMSE (BN-RMSE) within 
regions obtained by extracting the inner boundary of the annotation mask, dilating it using a $3\times3\times3$ structuring element for three iterations, and restricting the resulting neighborhood to the annotated region. 
For comprehensive statistical validation, two-sided Wilcoxon signed-rank tests were performed between NBGL and each baseline model using paired volume-level BN-RMSE values, with the Holm-Bonferroni correction applied across all pairwise comparisons. 
The resulting adjusted $p$-values are reported as $p_{\text{Holm}}$ in Table~\ref{tab:bn_rmse_evaluated}. Level-specific BN-RMSE results for each evaluated and unseen noise condition are provided in Supplementary Tables SIV and SV.

For the six evaluated noise levels, NBGL achieves the lowest local reconstruction errors in five configurations ($\sigma^2 \in \{0.005, 0.01, 0.02, 0.05, 0.1\}$). Under the most severe evaluated noise level ($\sigma^2 $=$0.2$), DESD-GAN obtains a marginally lower BN-RMSE than NBGL ($0.0825$ vs. $0.0833$), yielding a negligible difference of $0.0008$. A similar trend is observed across the unseen noise spectrum; NBGL claims the top performance under light ($\sigma^2$=$0.003$) and moderate ($\sigma^2$=$0.03$) conditions, securing BN-RMSE values of $0.0273$ and $0.0563$, respectively. Under extreme out-of-distribution noise ($\sigma^2$=$0.3$), DESD-GAN again exhibits a minute advantage in terms of absolute intensity deviation ($0.0884$ vs. $0.0889$, $\Delta$=$0.0005$). 

This subtle inflection at maximum noise scales can be attributed to adversarial generative baselines like DESD-GAN, which tend to synthesize highly smoothed intensity patches under heavy corruption, thereby favoring pixel-wise Euclidean distance metrics such as RMSE. However, as demonstrated in Table~\ref{tab:bn_rmse_evaluated} and the boundary pixel alignments in Fig.~\ref{fig:boundary_preservation_analysis}, this minor pixel-level compromise in NBGL allows our framework to prevent excessive smoothing and preserve the macroscopic geometric topology of the uterine cavity.

When aggregated across the entire spectrum, the superior capability of our model becomes evident. 
As summarized in Table~\ref{tab:bn_rmse_evaluated}, NBGL claims the lowest overall BN-RMSE across all evaluated levels ($0.0575$$\pm$$0.0078$) and all unseen levels ($0.0575$$\pm$$0.0076$), followed by DESD-GAN and DU-GAN. The Holm-adjusted $p$-values ($p_{\text{Holm}}$=$8.94 \times 10^{-8}$ against the majority of competitors, and $5.22 \times 10^{-7}$ against DESD-GAN on evaluated levels) confirm that the localized boundary-neighborhood refinement brought by NBGL is statistically significant and robust against volatile ultrasound noise variations.

\begin{table}[!t]
    \centering
    \caption{Aggregate BN-RMSE comparison of the proposed NBGL against 12 baseline models on the UterUS test set across the six evaluated noise levels and the three unseen noise levels. The best results are highlighted in bold; $p_{\mathrm{Holm}}$ denotes Holm-adjusted $p$-values.}
    \label{tab:bn_rmse_evaluated}
    \renewcommand{\arraystretch}{0.7}    
    \setlength{\tabcolsep}{2pt}
    \resizebox{\linewidth}{!}{%
    \begin{tabular}{l|cc|cc}
        \toprule
        \multirow{2}{*}{Method}
        & \multicolumn{2}{c|}{All evaluated levels}
        & \multicolumn{2}{c}{All unseen levels} \\
        \cmidrule(lr){2-3} \cmidrule(lr){4-5}
        & BN-RMSE$\downarrow$ & $p_{\mathrm{Holm}}$
        & BN-RMSE$\downarrow$ & $p_{\mathrm{Holm}}$ \\
        \midrule
        Gaussian Filter
        & {0.0692}{$\pm$0.0096} 
        & {8.94$\times 10^{-8}$} 
        & {0.0758}{$\pm$0.0106} 
        & {8.94$\times 10^{-8}$} \\
        Mean Filter
        & {0.0831}{$\pm$0.0139} 
        & {8.94$\times 10^{-8}$}
        & {0.0873}{$\pm$0.0144} 
        & {8.94$\times 10^{-8}$} \\
        Median Filter
        & {0.0887}{$\pm$0.0132} 
        & {8.94$\times 10^{-8}$}
        & {0.0970}{$\pm$0.0134} 
        & {8.94$\times 10^{-8}$} \\
        Bilateral Filter
        & {0.0960}{$\pm$0.0116} 
        & {8.94$\times 10^{-8}$}
        & {0.1121}{$\pm$0.0117} 
        & {8.94$\times 10^{-8}$} \\
        BM4D
        & {0.1085}{$\pm$0.0140} 
        & {8.94$\times 10^{-8}$}
        & {0.1199}{$\pm$0.0138} 
        & {8.94$\times 10^{-8}$} \\
        \midrule
        DESD-GAN
        & {0.0604}{$\pm$0.0095} 
        & {5.22$\times 10^{-7}$} 
        & {0.0611}{$\pm$0.0097} 
        & {8.94$\times 10^{-8}$} \\
        DU-GAN
        & {0.0662}{$\pm$0.0079} 
        & {8.94$\times 10^{-8}$}
        & {0.0683}{$\pm$0.0086} 
        & {8.94$\times 10^{-8}$} \\
        G2CR-FPN
        & {0.0711}{$\pm$0.0104} 
        & {8.94$\times 10^{-8}$}
        & {0.0738}{$\pm$0.0107} 
        & {8.94$\times 10^{-8}$} \\
        LIT-Former
        & {0.2116}{$\pm$0.0546} 
        & {8.94$\times 10^{-8}$}
        & {0.2109}{$\pm$0.0552} 
        & {8.94$\times 10^{-8}$} \\
        Speckle2Self
        & {0.1423}{$\pm$0.0486} 
        & {8.94$\times 10^{-8}$} 
        & {0.1420}{$\pm$0.0484} 
        & {8.94$\times 10^{-8}$} \\
        Autoencoder-KL
        & {0.1175}{$\pm$0.0192} 
        & {8.94$\times 10^{-8}$}
        & {0.1195}{$\pm$0.0204} 
        & {8.94$\times 10^{-8}$} \\
        DDIM
        & {0.2280}{$\pm$0.1060} 
        & {8.94$\times 10^{-8}$}
        & {0.2287}{$\pm$0.1049} 
        & {8.94$\times 10^{-8}$} \\
        \midrule
        \textbf{NBGL (Ours)}
        & \textbf{{0.0575}{$\pm$0.0078}} & \textbf{--} 
        & \textbf{{0.0575}{$\pm$0.0076}} & \textbf{--} \\
        \bottomrule
    \end{tabular}
    }
\end{table}


\subsection{Evaluation on External Data}
To further assess the applicability of NBGL to a different anatomical region and acquisition setting, additional experiments were conducted on the TRUSTED dataset~\citep{ndzimbong2025trusted}, a transabdominal kidney ultrasound dataset with expert anatomical landmark annotations, using a subject-level data split and the same mixed-speckle noise simulation settings (\ie, six speckle noise levels) as those in the main experiments. Detailed descriptions of the TRUSTED dataset and its data partition, along with the complete results for all baselines, are provided in the \emph{Supplementary Materials}. 
The experimental results are reported in Fig.~\ref{fig:trusted_quantitative}. 
This figure demonstrates that NBGL outperforms the competing methods, 
yielding higher mean PSNR and lower mean RMSE. More importantly, NBGL achieves the highest SSIM, indicating that the proposed framework  preserves structural information effectively in transabdominal kidney ultrasound, rather than only reducing intensity-level errors. 
In addition, the lowest BN-RMSE suggests better preservation of anatomical structures around the annotated landmark neighborhoods. 
Together with the relatively small standard deviations across the six noise levels, these results show that NBGL can be effectively applied to a different anatomical region and acquisition setting under heterogeneous speckle noise conditions.

\subsection{Limitations and Future Work}
Several limitations of the current study need to be considered. 
\emph{First}, the evaluation is based on controlled synthetic speckle simulation. Although the simulation follows a mean-preserving multiplicative model to approximate ultrasound speckle,  further validation on real clinical ultrasound data is needed to assess generalization to naturally occurring speckle patterns and acquisition-dependent noise characteristics. \emph{Second}, although the TRUSTED experiments demonstrate the applicability of NBGL to another anatomical region and acquisition setting, the current evaluation remains limited by the scale of available datasets. Future studies should include larger multi-center cohorts covering a broader range of anatomical targets. \emph{Third}, the evaluation mainly focuses on image quality and boundary-neighborhood metrics, while the influence of speckle reduction on downstream tasks remains to be further investigated. Future work will therefore focus on validation with real clinical data, extension to larger and more diverse ultrasound cohorts, and joint evaluation with downstream tasks such as anatomical structure segmentation and landmark localization.

\section{Conclusion}
\label{sec:conclusion}
This work presents NBGL, a noise-aware boundary-enhanced generative learning framework for ultrasound speckle reduction under varying speckle noise levels. NBGL integrates a speckle reduction branch and a boundary enhancement branch, where the former suppresses speckle noise and the latter learns boundary-sensitive representations from annotation-derived anatomical boundary supervision to mitigate over-smoothing. To adapt cross-branch feature coupling to heterogeneous noise levels, NBGL further incorporates NIWG and wFiLM. NIWG estimates the input speckle noise level using 3D Laplacian filtering and a median absolute deviation estimator, and maps the estimate to an interaction weight. wFiLM then uses this weight to modulate bidirectional feature coupling between the speckle reduction and boundary enhancement branches. 
Extensive evaluations on the UterUS dataset show that NBGL improves speckle reduction, structural preservation, and annotated-boundary consistency across multiple noise levels compared with state-of-the-art methods.

\newcommand{\pair}[2]{\makebox[3.8em][r]{$#1$}\,vs.\,\makebox[3.8em][l]{$#2$}}

\if false
\section*{Acknowledgments}
\fi

\bibliographystyle{model2-names}
\bibliography{reference}

\end{document}